\ifdefined\XeTeXversion\else\ifdefined\pdfoutput\pdfoutput=1\fi\fi
\documentclass[10pt,twocolumn,letterpaper]{article}
\usepackage[pagenumbers]{cvpr}

\usepackage{cuted} 
\usepackage{currfile} 
\usepackage{caption} 

\usepackage{microtype}

\renewcommand{\paragraph}[1]{\vspace{.5em}\noindent\textbf{#1.}} 

\usepackage[T1]{fontenc}
\usepackage{graphicx}
\usepackage{amsmath}
\usepackage{amssymb}
\usepackage{booktabs}
\usepackage{xcolor}
\newcommand{\best}[1]{\textbf{#1}}
\newcommand{\second}[1]{\underline{#1}}
\newcommand{\our}{FACET}
\let\method\our
\newcommand{\methodfull}{\textbf{F}lat \textbf{A}nisotropic \textbf{C}omponents for \textbf{E}diting and \textbf{T}racing}

\usepackage{multirow}
\usepackage{placeins}
\usepackage{float}
\definecolor{cvprblue}{rgb}{0.21,0.49,0.74}
\usepackage[pagebackref,breaklinks,colorlinks,allcolors=cvprblue]{hyperref}
\newif\ifarxiv
\arxivtrue

\title{\our{}: Ray-Traceable Local Implicit Neural Representations for Editable Images\protect\footnotemark}
\author{Rafa\l{} Tobiasz\textsuperscript{1,3} \quad Grzegorz Wilczy\'nski\textsuperscript{1,3} \quad Miko\l{}aj Zieli\'nski\textsuperscript{1,2,4}\\
Krzysztof Byrski\textsuperscript{1} \quad Dominik Belter\textsuperscript{2} \quad Przemys\l{}aw Spurek\textsuperscript{1,3}\\[0.5ex]
\textsuperscript{1}Jagiellonian University \qquad \textsuperscript{2}Pozna\'n University of Technology\\[0.2em]
\textsuperscript{3}IDEAS Research Institute \qquad \textsuperscript{4}Jagiellonian Center for Artificial Intelligence}

\begin{document}
\twocolumn[{\renewcommand\twocolumn[1][]{#1}\maketitle
\begin{center}
\centering
\includegraphics[width=\textwidth]{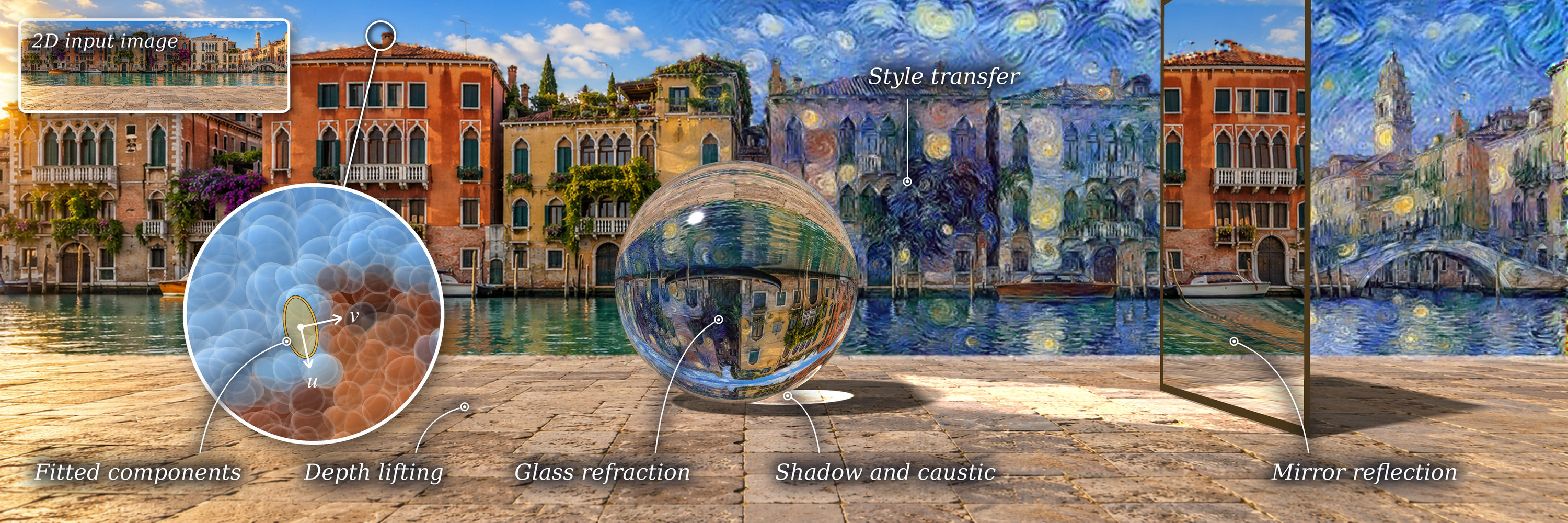}
\captionsetup{hypcap=false}
\captionof{figure}{\our{} brings 3D scene interactions to a single 2D image. Fitted planar components are lifted into depth, supporting appearance stylization and composition with glass and mirror objects through ray tracing. Reflections, refractions, shadows, and caustics remain consistent with the edited appearance. The magnifier shows the fitted components and their local coordinates.}
\label{fig:teaser}
\end{center}
}]
\ifarxiv{\renewcommand{\thefootnote}{\fnsymbol{footnote}}\footnotetext[1]{Code: \url{https://github.com/rtobiasz/FACET}}}\fi

\begin{abstract}
Implicit neural representations (INRs) encode images compactly as continuous functions, but they expose no surfaces that could be moved, lifted to 3D, or lit. Gaussian-based representations expose such surfaces. Some fix appearance to each primitive, so detail costs many primitives or memory. Others compute it from neighboring primitives, so edits change the appearance. Existing image representations cannot at once be deformed by hand or by physics simulation, be lifted into 3D, and be placed in a scene with added lights, mirrors, and glass, while keeping their fitted appearance. We introduce \our{}, an editable representation that brings 3D scene interactions to a single image. \our{} couples oriented planar components with continuous local neural appearance. A shared decoder predicts color and opacity from local coordinates and compact component descriptors, allowing fitted detail to follow geometry without per-component texture maps. Planar editing, mesh-driven simulation, depth-guided lifting, and ray tracing with reflection and refraction reuse the fitted appearance without retraining. Experiments on Kodak-24 and DIV2K-100 demonstrate higher reconstruction quality and smaller inference models than the evaluated editing-oriented baselines. Controlled editing and lifting experiments further demonstrate improved appearance preservation.

\end{abstract}

\begin{figure*}[!t]
\centering
\includegraphics[width=\textwidth]{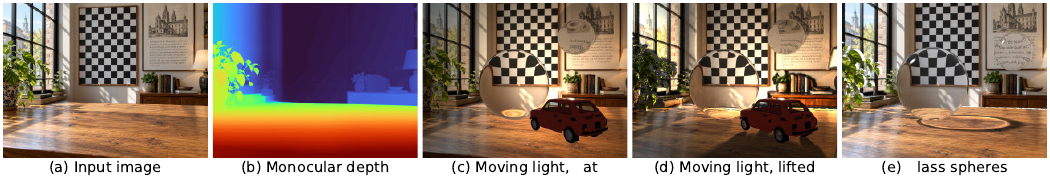}
\caption{One fitted image, lifted with monocular depth and ray-traced. An inserted car mesh shares the scene with the fitted image and casts shadows on it (c, d). Under the same moving light, lifting introduces depth-dependent shadows and occlusions. Inserted glass refracts the fitted scene.}
\label{fig:raytrace}
\end{figure*}

\section{Introduction}
\label{sec:introduction}

An image used in geometric editing must do more than reproduce its original pixels. Its content should bend with an edit, follow a physical simulation, and enter a three-dimensional scene where shadows, reflections, and refractions respond to geometry, as in Figs.~\ref{fig:teaser} and~\ref{fig:raytrace}. These operations require the fitted appearance to stay attached to the geometry that moves. Implicit neural representations (INRs)~\cite{sitzmann2020siren,tancik2020fourier,saragadam2023wire,liu2024finer,mueller2022instant} offer continuous, compact image fits, but edits must remap their global image coordinates, and the image field supplies no geometry to lift or ray trace. Explicit Gaussians~\cite{zhang2024gaussianimage,zhang2025imagegs,waczynska2025mirage} are movable, but their constant colors require many primitives for detail. Per-primitive textures~\cite{chao2025textured,svitov2025bbsplat} add variation while storing a map for each component. Decoding color from features aggregated over nearby points~\cite{jakubowska2026gainer} instead ties appearance to neighborhoods that an edit can change. As a result, none of them lets a single fitted image be deformed by hand or by physics simulation, lifted into depth, and ray traced with added lights, mirrors, and glass.

We introduce \our{} (\methodfull{}), which brings 3D scene interactions to a single image by combining explicit planar components with continuous local appearance. Unlike a global implicit field, each component can be manipulated directly, and its local coordinates capture spatially varying content rather than one color. One decoder shared by all components evaluates this function from local coordinates and a compact per-component descriptor, so detail is stored in shared weights rather than per-component textures, and a component's appearance does not depend on its neighbors. Geometry and appearance are thus coupled locally: component transformations change where image content is placed, while its appearance remains defined in the component's coordinate system. To support spatial interactions, we lift the fitted components into depth and render them through ray--component intersections. At each intersection, the local appearance function is evaluated at the corresponding coordinates, allowing the same image content to be queried along both camera rays and secondary rays. The lifted representation can therefore be composed with conventional mesh objects, supporting reflections, refractions, shadows, and caustics. These operations also remain connected to appearance editing: stylizing the component appearance changes what is seen directly, through glass, and in mirrors. \our{} thereby combines a continuous image representation, explicit geometric control, and ray-based scene composition within one model, allowing a fitted 2D image to participate in spatial effects while remaining editable.

Our contributions are:
\begin{itemize}
    \item \our{}, a local INR with a shared decoder predicting spatially varying color and opacity in movable planar component frames.
    \item Frame transport and depth-guided lifting enabling planar editing, mesh-driven simulation, and recursive ray tracing without refitting appearance.
    \item Higher reconstruction accuracy with a smaller inference state, better appearance preservation after smudge editing, and higher full-image PSNR after lifting than the evaluated baselines.
\end{itemize}

\begin{figure*}[!t]
\centering
\includegraphics[width=\textwidth]{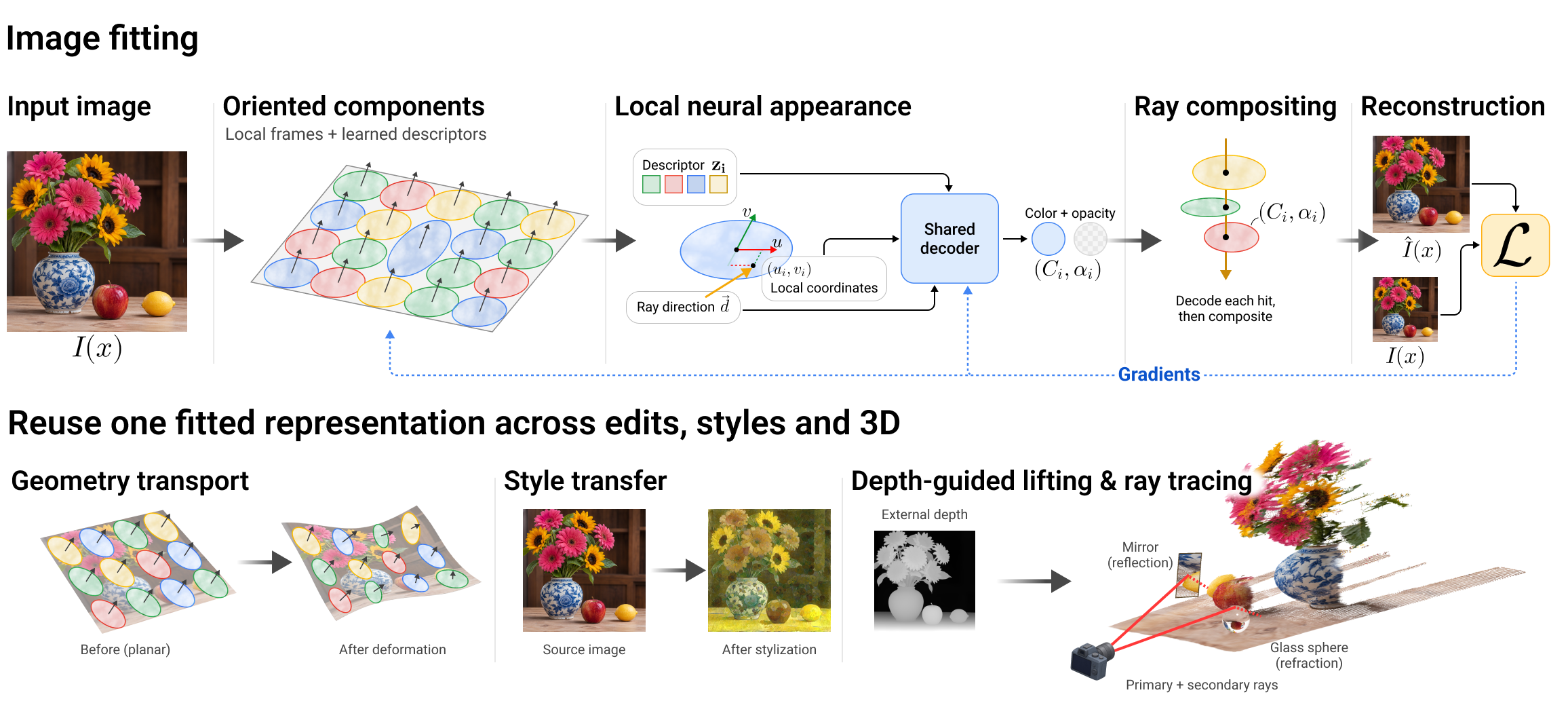}
\caption{\textbf{\our{} overview.} Fitting - The image is represented by oriented planar components, each with a local frame and a learned descriptor $\mathbf z_i$. At every ray hit, a shared decoder maps the local coordinates $\mathbf u_i$, the descriptor, and the fixed ray direction $\mathbf d$ to color and opacity $(\mathbf C_i,\alpha_i)$. Hits are composited in ray order, and the image loss jointly optimizes components and decoder. Reuse - Deformations transport each component's frame, so its fitted pattern moves with it. Stylization optimizes appearance while geometry stays fixed. Lifting with an external depth map places the components in 3D, where reflected and refracted rays from inserted objects query the same decoder.}
\label{fig:method}
\end{figure*}

\section{Related Work}
\label{sec:related}
\paragraph{Continuous image representations}
Coordinate networks capture image detail through Fourier features~\cite{tancik2020fourier} or periodic, wavelet, or adaptive activations, as in SIREN~\cite{sitzmann2020siren}, WIRE~\cite{saragadam2023wire}, and FINER~\cite{liu2024finer}. Multiresolution hash encodings in Instant-NGP~\cite{mueller2022instant} and adaptive radial bases in NeuRBF~\cite{chen2023neurbf} offer other continuous parameterizations. These compact image fields index appearance by global coordinates. Geometric edits must invert or remap that coordinate map, and the fields provide no explicit geometry for lifting or secondary-ray intersections.

\begin{figure}[!tb]
\centering
\includegraphics[width=\columnwidth]{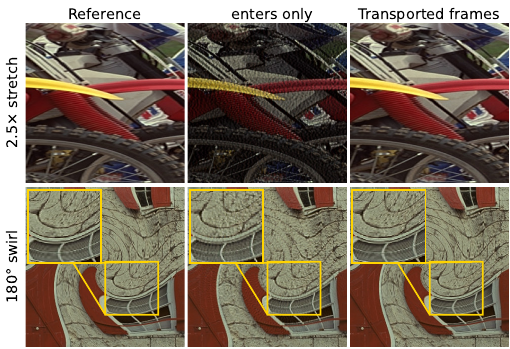}
\caption{Stretch and $180^\circ$ swirl on crops of Kodak images: reference, moved centers only, and transported frames. Transport closes the gaps under stretch and rotates local appearance with the swirl.}
\label{fig:transport}
\end{figure}

\begin{figure*}[!t]
\begin{minipage}[t]{\columnwidth}
\centering
\includegraphics[width=\columnwidth]{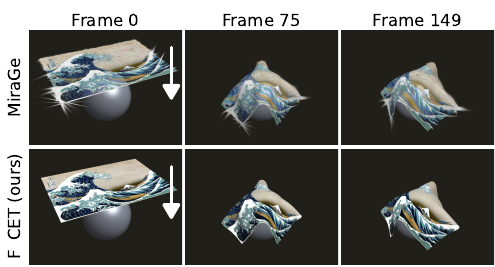}
\caption{Cloth renders on the same mesh and simulation: The cloth falls onto a sphere (arrow). Oblique views expose boundary spikes already before motion and washed-out colors on MiraGe's folds. \our{} carries cleaner fitted appearance through the folds.}
\label{fig:cloth}
\end{minipage}
\hfill
\begin{minipage}[t]{\columnwidth}
\centering
\includegraphics[width=\columnwidth]{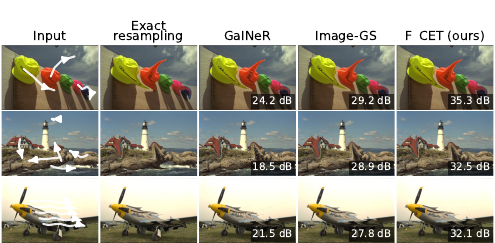}
\caption{Smudge edits with shared warps. Exact resampling is the bicubic inverse warp of the input. Image-GS centers follow the warp and its covariances follow the Jacobian, as for \our{}'s components. White arrows mark brush paths, and scores measure PSNR on brushed pixels.}
\label{fig:smudge}
\end{minipage}
\end{figure*}

\paragraph{Gaussian image representations}
Gaussian image methods prioritize reconstruction, compression, and fitting speed. GaussianImage~\cite{zhang2024gaussianimage} and GaussianImage++~\cite{li2026gaussianimagepp} optimize image representation and compression. Image-GS~\cite{zhang2025imagegs} and Large Images Are Gaussians~\cite{zhu2025lig} use hierarchies to recover coarse structure and fine detail. Instant GaussianImage~\cite{zeng2025instantgi} amortizes initialization across images. SGI~\cite{pan2026sgi} decodes Gaussians from structured seeds and small MLPs for compact large-image storage. Their constant-color Gaussians require many primitives for detail and carry no spatial color variation within a deformed primitive. \our{} instead decodes spatially varying color inside each component, so detail moves with the component.

\paragraph{Editable image representations}
Image vectorization fits editable shapes to photographs with texture-encapsulated parameters~\cite{chen2024tesp}, multi-layer features~\cite{chen2025easyeditable}, or illumination-aware layers~\cite{lin2026covec}, but targets 2D editing rather than simulation, lifting, or ray queries.
MiraGe~\cite{waczynska2025mirage} edits flat, constant-color Gaussians fitted from one view through geometric and physics-driven manipulation. GaINeR~\cite{jakubowska2026gainer} decodes color from features aggregated over nearby Gaussians and demonstrates edits, simulation, and lifting. These are our main image-editing baselines. Their appearance is either constant within a primitive or depends on a neighborhood that deformation can change, leaving compact detail attached to each moving primitive unresolved. \our{} conditions its decoder on the component's own descriptor and local coordinates, so moving other components does not change its decoded appearance.

\paragraph{Appearance within Gaussian primitives}
3DGS~\cite{kerbl2023gaussians} represents a scene with volumetric Gaussians, while 2DGS~\cite{huang2024gs2d} uses oriented planar disks with perspective-correct ray--splat intersections. Textured Gaussians~\cite{chao2025textured} attach RGB and alpha texture maps to Gaussian supports. BBSplat~\cite{svitov2025bbsplat} uses textured planar billboards with learnable RGB and alpha maps, including ray-tracing effects. GStex~\cite{rong2025gstex} assigns RGB texture maps to 2D Gaussians to separate appearance detail from geometry and enable retexturing. Neural Texture Splatting~\cite{wang2025neuraltexture} predicts per-primitive local appearance by querying a global neural field at each primitive's position, for multi-view reconstruction. Neural Shell Texture Splatting~\cite{zhang2025neuralshell} queries a shared scene-space texture at ray--primitive intersections. \our{} works on a single image: one shared network, conditioned on a compact descriptor that each component carries, decodes each component in its own local frame, so moving the frame carries fitted appearance without storing or resampling a per-component texture.

\paragraph{Local neural fields and ray queries}
Scaffold-GS~\cite{lu2024scaffold} decodes Gaussians from anchor features for view-adaptive rendering. EKS~\cite{zielinski2026eks} and IRIS~\cite{wilczynski2026iris} use Gaussian anchors for multi-view 3D editing. REdiSplats~\cite{byrski2025redisplats} edits flat Gaussians, while 3DGRT~\cite{moenneloccoz2024grt} supports secondary rays through volumetric Gaussians. PhysGaussian~\cite{xie2024physgaussian} transports Gaussian geometry using the material point method (MPM)~\cite{stomakhin2013mpm}. Layered depth images~\cite{shade1998ldi} store occluded surfaces, and single-image 3D photography~\cite{shih2020photography,tucker2020mpi} inpaints hidden content, while \our{} does not. These methods establish geometric manipulation and ray queries, but do not address carrying appearance fitted to a single image through both edits and depth-guided lifting.

\paragraph{Stylization} Neural style transfer~\cite{gatys2016style} and CLIPstyler~\cite{kwon2022clipstyler} restyle pixels. CLIPGaussian~\cite{howil2025clipgaussian} and OmniStyle-INR~\cite{kajca2026omnistyle} optimize Gaussian or implicit appearance using multimodal style objectives. For geometric reuse, restyled appearance must also stay attached to movable geometry. We use their style sets to evaluate this combination.

\section{Method}
\label{sec:method}
\our{} attaches a small neural appearance field to each oriented planar component. Fig.~\ref{fig:method} summarizes the pipeline. During fitting, each ray hit is expressed in a component's local frame, decoded into color and opacity, and blended with the other hits to reconstruct the source image. The image loss jointly optimizes component geometry, appearance parameters, and the shared decoder. After fitting, geometric operations move the components and their local frames while retaining the learned appearance.

\begin{figure*}[!t]
\centering
\includegraphics[width=\textwidth]{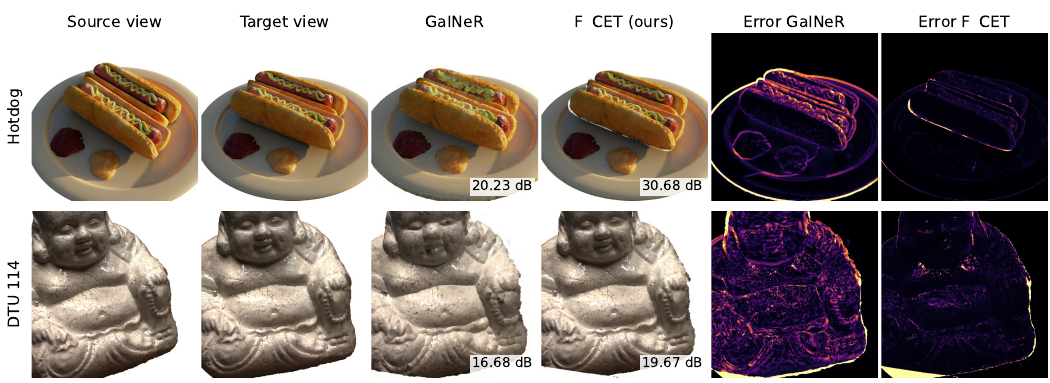}
\caption{Single-image lifting on Hotdog and DTU 114: source, target, GaINeR, \our{}, and mean absolute RGB-error maps on a shared scale. Panels use the same GT-depth renders as Table~\ref{tab:lift}. Tile scores are per-view PSNR.}
\label{fig:lift}
\end{figure*}

\subsection{Local neural image representation}
\label{sec:representation}
We center an image of width $W$ and height $H$ at the origin and scale its coordinates so the width spans two units, preserving its aspect ratio. Pixel spacing is $\Delta=2/W$, and orthographic fitting rays pass through the pixel centers, perpendicular to the plane. Rendering at another resolution samples the same image region on a different grid. We represent the image with a set of components. Component $i$ has center $\boldsymbol\mu_i\in\mathbb R^2$, orientation $\theta_i\in\mathbb R$, log-scales $\mathbf s_i\in\mathbb R^2$, and support shape parameter $k_i\in\mathbb R$. Its appearance is parameterized by base color $\mathbf c_i\in\mathbb R^3$, opacity logit $a_i\in\mathbb R$, and a learned descriptor $\mathbf z_i\in\mathbb R^4$. A ray meets the image plane at $\mathbf x$. In component $i$'s frame, this point has local coordinates $\mathbf u_i$ and support function $g_i$:
\begin{align}
\mathbf u_i &= \operatorname{diag}(e^{-\mathbf s_i})\mathbf R(\theta_i)^\top
(\mathbf x-\boldsymbol\mu_i), \\
 g_i &= \exp\!\left[-\frac{\|\mathbf u_i\|^{2\kappa_i}}{2\kappa_i}\right],
\quad \kappa_i=1+\operatorname{softplus}(k_i).
\label{eq:primitive}
\end{align}
Here $\mathbf R(\theta_i)$ is the 2D rotation by $\theta_i$, and the local coordinates $\mathbf u_i=(u,v)$ are measured in units of the component's scales. The support function $g_i$ is a generalized Gaussian. Its support is the elliptical region of the plane where $g_i$ is non-negligible, with shape set by the scales, angle, and $\kappa_i$. The edge becomes sharper as $\kappa_i$ grows.
At each ray intersection, the shared appearance decoder returns four values $D_\phi(\mathbf u_i,\mathbf z_i,\mathbf d)=(\mathbf q_i,q_i^\alpha)$, where $\mathbf d$ is the ray direction. The decoder uses a learned four-level grid encoding of local coordinates, two hidden layers of width 64 with hard-swish activations, and four outputs. Its encoding and weights are shared by all components and negligible in size relative to them. The ray direction is constant during single-image fitting and held at that direction after lifting, so the decoded appearance is view-independent. With a single direction, this input acts as a learned bias. The decoded RGB values modulate the base color, while decoded opacity modulates the spatial support:
$\mathbf C_i=\mathbf c_i\odot\operatorname{ReLU}(\mathbf q_i)$ and $\alpha_i=\sigma(a_i)g_i\sigma(q_i^\alpha)$. For hits $i_1,\dots,i_m$ in ray order, front-to-back alpha compositing~\cite{porter1984compositing} gives the image color
\begin{equation}
\widehat I(\mathbf x)=\sum_j T_j\alpha_{i_j}\mathbf C_{i_j}+T_{m+1}\mathbf b,
\quad T_j=\prod_{\ell<j}(1-\alpha_{i_\ell}),
\label{eq:render}
\end{equation}
where $\mathbf b$ is the background and $T_j$ is the fraction of light transmitted through earlier hits. All components lie in one plane, so a ray meets them at the same depth and geometry does not define their compositing order. We composite them in a fixed order given by their index, and the fit adapts to this order.

\paragraph{Fitting a coplanar field}\label{sec:fitting}
We randomly initialize a fixed set of coplanar components and jointly optimize their geometry, appearance parameters, and shared decoder without densification. The default model fits 500k components for 30k iterations and stores 14 float32 scalars per component, giving a 28\,MB inference state including the shared decoder. The source image supplies all supervision. The loss combines mean squared RGB error between clamped renders and normalized input pixels with scale regularization. At the default budget, component scales are bounded above by about one pixel ($\Delta$) and below by half of that bound.

Rendered colors are clamped to the valid range, and out-of-range channels receive no image-loss gradient. We let the gradient pass through the clamp only when it points back into the valid range, which stabilizes fitting without changing the rendered image.

\subsection{Transport under planar deformations}
\label{sec:transport}
An edit moves component centers, but center motion alone leaves their supports and local appearance frames unchanged (Fig.~\ref{fig:transport}). We also apply the local deformation to each frame. We call this \emph{frame transport}: the component's support and appearance frame move, rotate, and stretch with the local deformation. For a deformation that is locally $\mathbf x'-\boldsymbol\mu_i'=\mathbf F_i(\mathbf x-\boldsymbol\mu_i)$ around the component, with $\det\mathbf F_i>0$, the full transported frame gives
\begin{equation}
\begin{aligned}
\mathbf A_i&=\mathbf F_i\mathbf R(\theta_i)\operatorname{diag}(e^{\mathbf s_i}),\\
\mathbf u_i'&=\mathbf A_i^{-1}(\mathbf x'-\boldsymbol\mu_i')=\mathbf A_i^{-1}\mathbf F_i(\mathbf x-\boldsymbol\mu_i)=\mathbf u_i.
\end{aligned}
\label{eq:transport}
\end{equation}
The decoder receives the same local coordinates and descriptor, preserving appearance under this exact affine transport.
A singular value decomposition (SVD) of $\mathbf A_i$ gives the axes and scales of the deformed ellipse. Singular axes have sign and permutation ambiguities that would mirror or swap the decoder's $uv$ frame. Among equivalent proper frames, we select the one nearest the original frame rotated by the polar factor of $\mathbf F_i$. Our two-scale, one-angle ellipse frame realizes this invariance exactly for similarity transforms, but only approximately under anisotropic stretch and shear. Querying the decoder through the exact local affine map changes PSNR by at most 0.14\,dB on our warp tests. In elastic MPM simulations, $\mathbf F_i$ comes from the solver. For sand-like plastic flow we estimate a regularized total deformation from neighboring primitive centers, because the solver's elastic factor omits plastic motion.

\begin{figure*}[!t]
\centering
\includegraphics[width=\textwidth]{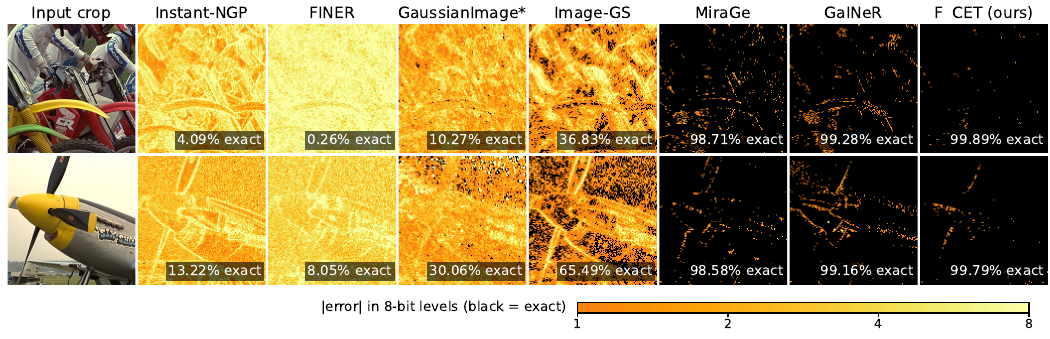}
\caption{Local reconstruction errors on Kodak images 5 and 20. The color scale spans 1--8 eight-bit levels (black is exact, errors of 8 or more saturate). Each displayed cell is the maximum error over a $2\times2$ pixel block so isolated errors remain visible at print size. Labels give unchanged full-image exact RGB-pixel fractions. GaussianImage$^*$ as in Table~\ref{tab:reconstruction}.}
\label{fig:recon-errors}
\end{figure*}

\begin{table*}[!t]
\centering\small
\setlength{\tabcolsep}{6pt}
\renewcommand{\arraystretch}{1.08}
\caption{Reconstruction quality and cost on one RTX 4090. PSNR is the mean float PSNR over three seeds. DIV2K-100 uses the validation images downsampled by two. Training time is a per-image mean on Kodak-24 and DIV2K images 0801--0810. Mem. is the uncompressed inference state. \our{}, GaINeR, MiraGe, and Image-GS start from 500k primitives. Training times use the optimized kernel, which matches the reference kernel's Kodak quality. FPS is a static native-size render averaged over three images per dataset. Bold marks the best and underline the second-best value within each group. $^*$70k Gaussians, 50k steps. $^\dagger$ $2^{19}$-entry hash table.}
\label{tab:reconstruction}
\begin{tabular}{clcccccccc}
\toprule
 & \textbf{Method} & \multicolumn{2}{c}{\textbf{PSNR (dB)}} & \multicolumn{2}{c}{\textbf{Train (min)}} & \multicolumn{2}{c}{\textbf{Mem. (MB)}} & \multicolumn{2}{c}{\textbf{FPS}}\\
\cmidrule(lr){3-4}\cmidrule(lr){5-6}\cmidrule(lr){7-8}\cmidrule(lr){9-10}
 & & \textbf{Kodak} & \textbf{DIV2K-100} & \textbf{Kodak} & \textbf{DIV2K} & \textbf{Kodak} & \textbf{DIV2K} & \textbf{Kodak} & \textbf{DIV2K}\\
\midrule
\multirow{4}{*}{\rotatebox[origin=c]{90}{\scriptsize\shortstack{Coordinate\\networks}}} & SIREN~\cite{sitzmann2020siren} & 37.58 & 33.42 & 6.3 & 11.2 & \best{0.8} & \best{0.8} & \second{65} & \second{40}\\
 & WIRE~\cite{saragadam2023wire} & 32.00 & 29.92 & 10.7 & 19.0 & \best{0.8} & \best{0.8} & 22 & 13\\
 & FINER~\cite{liu2024finer} & \second{39.31} & \second{34.26} & \second{4.2} & \second{7.5} & \best{0.8} & \best{0.8} & 38 & 22\\
 & Instant-NGP$^\dagger$~\cite{mueller2022instant} & \best{44.27} & \best{41.91} & \best{0.6} & \best{0.6} & 30.7 & 30.7 & \best{4433} & \best{3953}\\
\midrule
\multirow{3}{*}{\rotatebox[origin=c]{90}{\scriptsize 2D GS}} & GaussianImage$^*$~\cite{zhang2024gaussianimage} & 44.03 & 39.52 & \second{0.9} & \second{1.0} & \best{2.2} & \best{2.2} & \best{2434} & \best{2349}\\
 & GaussianImage++~\cite{li2026gaussianimagepp} & \best{53.56} & \best{50.48} & \best{0.7} & \best{0.9} & \second{5.2} & \second{7.6} & \second{2022} & \second{1578}\\
 & Image-GS~\cite{zhang2025imagegs} & \second{50.37} & \second{48.58} & 2.5 & 2.8 & 16.0 & 16.0 & 402 & 400\\
\midrule
\multirow{3}{*}{\rotatebox[origin=c]{90}{\scriptsize\shortstack{Edit-\\oriented}}} & MiraGe~\cite{waczynska2025mirage} & 71.33 & \second{65.66} & \best{7.3} & \best{12.1} & 212.7 & 258.9 & \best{206} & \best{194}\\
 & GaINeR~\cite{jakubowska2026gainer} & \second{72.97} & 61.34 & 24.4 & 42.7 & \second{44.1} & \second{44.1} & 76 & 49\\
 & \textbf{\our{} (ours)} & \best{81.35} & \best{70.29} & \second{15.5} & \second{20.3} & \best{28.0} & \best{28.0} & \second{120} & \second{95}\\
\bottomrule
\end{tabular}
\end{table*}
\paragraph{Mesh-driven deformation} For mesh-cage deformation, we bind each component to a triangle of a regular image-plane grid. Barycentric coordinates move its center with the triangle, while interpolated vertex tangents transport its two local axes. An SVD restores a representable tangent frame. The fitted descriptor and decoder remain unchanged. In the cloth examples (Fig.~\ref{fig:cloth}), the cage surface provides the visibility depth and shading normal for ray tracing with inserted geometry. Fig.~\ref{fig:smudge} shows planar brush edits under the same transport.

\subsection{Depth-guided lifting and ray tracing}
\label{sec:lifting}
A depth map, from a monocular estimator such as Depth Pro~\cite{bochkovskii2025depthpro} or from ground truth when a benchmark provides it, places the fitted image in 3D (Fig.~\ref{fig:lift}). A component centered at source pixel $\mathbf p=(p_x,p_y)$ with depth $Z(\mathbf p)$ moves to
\begin{equation}
\mathbf P=Z(\mathbf p)\big[(p_x-c_x)/f_x,\;(p_y-c_y)/f_y,\;1\big]^\top,
\end{equation}
where $f_x,f_y$ are focal lengths and $(c_x,c_y)$ is the principal point. With $\mathbf r(\mathbf p)=\big[(p_x-c_x)/f_x,\;(p_y-c_y)/f_y,\;1\big]^\top$, the Jacobian of this back-projection is
\begin{equation}
\mathbf J=\frac{\partial\mathbf P}{\partial\mathbf p}=Z(\mathbf p)\begin{bmatrix}1/f_x&0\\0&1/f_y\\0&0\end{bmatrix}+\mathbf r(\mathbf p)\,\nabla Z(\mathbf p)^\top.
\end{equation}
The lifted axes are the columns of $\mathbf J\mathbf R(\theta_i)\operatorname{diag}(e^{\mathbf s_i})$, with the in-plane axes expressed in pixels. Their polar factor gives the 3D tangent frame and their lengths give the scales. Each lifted component acts as an oriented surface element~\cite{pfister2000surfels,zwicker2001surface}. The depth-gradient term tilts components on slanted surfaces, such as a table seen at an angle. Descriptors, base colors, base opacity, and the decoder are reused unchanged. Fits used for lifting draw a random background color in each iteration. With a fixed background, nearly transparent components can reproduce the image as well as opaque ones, so opacity is ambiguous. Random backgrounds remove this ambiguity, and lifted surfaces do not let the background show through.

Three rendering choices keep the fitted appearance after lifting. First, the fit learned its appearance under the fixed index order of coplanar components. After lifting, overlapping components lie at different depths and that order would change. We therefore sort hits by $t+\delta\,\rho\,i/2^{19}$, where $t$ is the hit distance, $\rho$ the hit's distance from the source camera, $i$ the component index, and $\delta=0.01$. Since indices stay below $2^{19}$, two hits can swap order only when their distances differ by less than 1\% of $\rho$. Separated surfaces keep their geometric order, while nearly coincident supports tend to follow the fitted source order. Without this offset, full-image PSNR of the lifted views drops by 1.2--2.6\,dB on the GT-depth scenes of Table~\ref{tab:lift}. Second, the decoder was fitted from a single view direction, so lifted views use that same direction. Third, depth-edge gradients use bounded minmod of one-sided differences with an $85^\circ$ slant threshold to avoid streaks. Primary and secondary rays, including reflections and refractions, can intersect the lifted components in a recursive ray tracer.

\begin{figure}[!tb]
\centering
\includegraphics[width=\columnwidth]{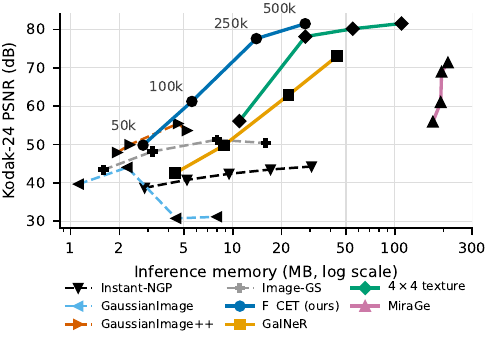}
\caption{Kodak-24 quality versus inference memory (log scale), single-seed. \our{} and GaINeR use 50k--500k components. The texture stores $4\times4$ RGB values per component. MiraGe starts from 50k--500k Gaussians, and densification brings every run to a similar size. Dashed lines show image-fitting methods without editing support at several budgets. Instant-NGP varies its hash table from $2^{15}$ to $2^{19}$ entries.}
\label{fig:capacity}
\end{figure}

\section{Experiments}
\label{sec:experiments}
We test three claims enabled by component-local appearance: accurate image fitting, appearance preserved through edits, and higher full-image PSNR after lifting than the evaluated baselines. Baselines use official code. PSNR is computed on unquantized renders against the 8-bit input scaled to $[0,1]$. All fitted methods report their best training-PSNR checkpoint, and memory denotes the uncompressed inference state. Exact fitting settings, per-method protocols, and further results, including MPM examples, are in Appendix~\ref{sec:supp_protocols}.

\subsection{Reconstruction and computational cost}
\label{sec:reconstruction}

\our{} reaches the highest reconstruction PSNR of all compared methods on both Kodak-24~\cite{franzen1999kodak} and DIV2K~\cite{agustsson2017div2k} (Table~\ref{tab:reconstruction} and Fig.~\ref{fig:recon-errors}). At the same 500k primitive budget as Image-GS, \our{} fits far more accurately because its decoder models spatial variation inside each component.

\our{} has the smallest inference state among the evaluated edit-oriented methods and trains and renders faster than GaINeR (Table~\ref{tab:reconstruction}). Sharing one decoder avoids storing detailed appearance separately in every component, while training uses less than a quarter of GaINeR's peak GPU memory on the measured Kodak image. MiraGe trains and renders faster but uses a far larger model. Rebuilding and rendering during edits takes about 10\,ms per frame. The 28\,MB inference state targets editing and ray tracing, not compression. Shared appearance thus makes accurate fits practical to edit.

\begin{table}[!tb]
\centering\small
\setlength{\tabcolsep}{4pt}
\caption{Single-seed component ablations on Kodak-24: mean float PSNR (dB) and change from full \our{}. Local appearance and coordinates account for the largest gains. Constant per-primitive color disables the decoder, while without local $uv$ the decoder sees only the descriptor.}
\label{tab:ablations}
\begin{tabular}{lrr}
\toprule
\textbf{Variant} & \textbf{PSNR} & \textbf{$\Delta$}\\
\midrule
\textbf{Full FACET} & \best{81.49} & --\\
Constant per-primitive color & 72.19 & $-$9.30\\
Without local $uv$ & 73.64 & $-$7.85\\
Without descriptor & 80.14 & $-$1.35\\
\bottomrule
\end{tabular}
\end{table}
\subsection{Capacity and component analysis}
\label{sec:analysis}
\our{} exceeds GaINeR on every image at every tested budget with less memory (Fig.~\ref{fig:capacity}). Every MiraGe fit uses more memory than the largest \our{} model while reconstructing worse, and a per-component texture needs several times more memory for comparable quality. The ablations in Table~\ref{tab:ablations} explain the reconstruction advantage in Table~\ref{tab:reconstruction}: constant color and removing local coordinates cause the largest losses. Sharing one decoder makes spatially varying appearance affordable without storing a texture per component.

\begin{table}[!tb]
\centering\footnotesize
\setlength{\tabcolsep}{2pt}
\caption{Full-image novel-view PSNR and uncovered target-object pixels (holes) after lifting one image with hole-filled ground-truth (GT) or Depth Pro (DP) depth. The textured mesh uses the same image and depth, removing triangles above \our{}'s $85^\circ$ limit.}
\label{tab:lift}
\begin{tabular}{llrrrrr}
\toprule
\textbf{Scene} & \textbf{Depth} & \multicolumn{3}{c}{\textbf{PSNR (dB)}} & \multicolumn{2}{c}{\textbf{Holes (\%)}}\\
\cmidrule(lr){3-5}\cmidrule(lr){6-7}
 & & \textbf{GaINeR} & \textbf{Mesh} & \textbf{\our{}} & \textbf{Mesh} & \textbf{\our{}}\\
\midrule
Hotdog & GT & 20.15 & 22.95 & \best{30.65} & 2.3 & \best{0.1}\\
Hotdog & DP & 19.29 & 18.85 & \best{20.68} & 3.7 & \best{0.7}\\
Lego & GT & 16.27 & 16.45 & \best{21.41} & 13.9 & \best{2.8}\\
Lego & DP & 15.65 & 14.66 & \best{15.78} & 15.8 & \best{8.6}\\
DTU 114 & GT & 17.31 & 19.16 & \best{20.55} & 2.9 & \best{1.6}\\
DTU 114 & DP & 16.48 & 18.55 & \best{19.04} & 3.1 & \best{1.7}\\
DTU 110 & GT & 13.91 & 15.14 & \best{16.74} & 5.7 & \best{2.3}\\
DTU 110 & DP & 13.43 & 18.45 & \best{19.16} & 2.3 & \best{0.6}\\
\bottomrule
\end{tabular}
\vspace{-12pt}
\end{table}
\subsection{Geometric manipulation and lifting}
\label{sec:manipulation}

\paragraph{Ellipse transport under controlled warps}\label{sec:transport-eval}
Transporting component frames outperforms moving only centers in every tested image--warp pair, most under stretch (Fig.~\ref{fig:transport}). Center motion leaves support sizes and orientations unchanged, opening gaps in stretched regions. Transporting the frame carries the local appearance with the deformation, as measured against bicubic~\cite{keys1981cubic} inverse warping. Under unit shear, updating rotation or scales alone falls well short of full transport. Rotation alone suffices for pure rotation, but both rotation and scale are needed to preserve appearance under more general warps.

\our{} has the best brushed-region fidelity, against a direct bicubic warp of the input (Fig.~\ref{fig:smudge}). The brush rearranges the neighborhoods on which GaINeR's appearance depends, whereas each \our{} component carries its own locally decoded appearance. Image-GS transports its geometry correctly but starts from a less accurate constant-color fit. \our{} wins every case against GaINeR across all Kodak images and stroke sets for each tested \our{} seed, and also wins every case at the smaller tested budgets, where the gap widens. At matched memory, a per-component texture loses to \our{} in 91 of 96 cases for the default seed. Local appearance and transported frames therefore preserve detail beyond what movable geometry alone provides.

\paragraph{Edits and simulation} One fitted representation handles geometric edits, cloth motion, and sand-like flow (Figs.~\ref{fig:edits} and~\ref{fig:cloth}). A mesh textured with the source image can handle cloth, but \our{} also works without a mesh and follows flow that breaks image connectivity. Transported frames carry each component's local appearance through these motions. Learned per-component opacity keeps thin silhouettes, while GaINeR's binary mask gives staircase edges. MiraGe's front-fitted flat Gaussians show spikes in oblique views and washed-out colors on folds.

\begin{table}[t]
\centering\small
\setlength{\tabcolsep}{3pt}
\caption{DIV2K $8\times$ super-resolution on 23 images. \emph{Operator} marks methods that use the bicubic degradation model.}
\label{tab:sr_main}
\begin{tabular}{lcccc}
\toprule
\textbf{Method} & \textbf{Operator} & \textbf{PSNR $\uparrow$} & \textbf{SSIM $\uparrow$} & \textbf{LPIPS $\downarrow$}\\
\midrule
Bicubic & no & 23.14 & .637 & .520\\
GaINeR & no & 23.16 & .636 & .520\\
MiraGe & no & 23.26 & .639 & .514\\
\midrule
IBP (20 steps) & yes & 23.55 & .648 & .501\\
\our{} & yes & \textbf{23.85} & \textbf{.658} & \textbf{.489}\\
\bottomrule
\end{tabular}
\vspace{-12pt}
\end{table}
\paragraph{Depth-guided lifting} \our{} has higher full-image PSNR than GaINeR and the textured mesh in every tested setting, with fewer holes than the mesh (Table~\ref{tab:lift} and Fig.~\ref{fig:lift}). These held-out views cover Hotdog, Lego~\cite{mildenhall2020nerf}, and DTU~\cite{jensen2014dtu} with GT or Depth Pro~\cite{bochkovskii2025depthpro} depth. Lifted components retain their local frames, descriptors, and shared decoder, so their fitted appearance is reused unchanged. Overlapping components cover small gaps that the mesh leaves at cut depth edges. Lego with Depth Pro is a near-tie with GaINeR, and the mesh has higher SSIM~\cite{wang2004ssim} or better LPIPS~\cite{zhang2018lpips} in some settings.

Lifted components also preserve fitted detail in reflections and refractions (Figs.~\ref{fig:teaser} and~\ref{fig:raytrace}). Rays from inserted objects~\cite{whitted1980illumination} query the same local decoder without refitting appearance. With an inserted glass ball, a $1024\times768$ frame renders in about 50\,ms with $2\times2$ supersampling on one RTX 4090. Source illumination stays baked in, so added lights cast shadows and caustics but materials are not recovered.

\subsection{Super-resolution and stylization}
\label{sec:applications}
\our{} improves over iterative back-projection~\cite{irani1991ibp} on every evaluated super-resolution image and has the best mean PSNR, SSIM, and LPIPS among the compared methods (Table~\ref{tab:sr_main}). Fitting continuous local appearance through the known bicubic degradation operator allows the shared decoder to sample a finer output grid. Stylization optimizes appearance with CLIP~\cite{radford2021clip} while keeping component geometry fixed (Fig.~\ref{fig:method}, bottom). On the image-style set of OmniStyle-INR~\cite{kajca2026omnistyle}, \our{} reaches a CLIP-based style similarity of 71.7, versus 70.0 for OmniStyle-INR and 66.4 for CLIPGaussian~\cite{howil2025clipgaussian}. With text prompts it matches CLIPGaussian's style similarity while preserving more content. Both applications retain the local appearance representation, so their results remain editable and can be deformed or lifted.

\FloatBarrier
\par\vspace{3pt}\begingroup
\centering
\includegraphics[width=\columnwidth]{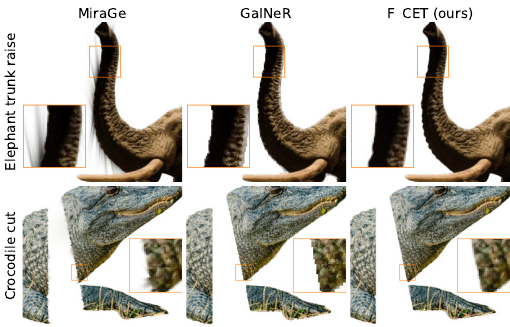}
\captionof{figure}{Geometric edits under the same warp and object mask for all methods. MiraGe shows spikes and blur, GaINeR's binary mask leaves staircasing and light fringes, and \our{} preserves smooth silhouettes. All four edits appear in Fig.~\ref{fig:supp_edits}.}
\label{fig:edits}
\par\endgroup
\section{Conclusion}
\label{sec:conclusion}
Continuous image fields lack movable geometry, while Gaussian appearance is constant, stored per primitive, or tied to neighbors. \our{} instead queries one shared decoder in each planar component's frame, keeping appearance attached through deformation, lifting, and ray tracing. It reconstructs best among compared methods on both datasets, with the smallest model among the edit-oriented methods, and better preserves appearance after smudging and lifting. The same fit supports simulation, reflections, refractions, and restyling, turning one image into a reusable asset beyond its pixel grid.

\paragraph{Limitations}
Depth-guided lifting depends on the supplied depth and, since \our{} is not generative, leaves holes where surfaces were hidden in the source image. Fitting takes minutes per image on one GPU, so \our{} targets preparing an image for later editing rather than fast encoding.

\clearpage
\appendix
\section*{Appendix}
\raggedbottom
\makeatletter
\setlength{\@fptop}{0pt}
\setlength{\@dblfptop}{0pt}
\makeatother
\section{Protocols and implementation}
\label{sec:supp_protocols}
\paragraph{Code} Configuration files for all reported results are provided at \url{https://github.com/rtobiasz/FACET}.

\paragraph{Image sources} The Venice input image of the main-text teaser (Fig.~\ref{fig:teaser}) was generated with an image-generation model (ChatGPT). The checkerboard-room image was edited with SDXL inpainting to remove a chair before fitting. All other inputs are photographs, paintings, or images from the evaluated datasets.

\paragraph{Model variants} RGBA and lifting fits randomize backgrounds (Tables~\ref{tab:smudge} and~\ref{tab:supp_lift_metrics}). Capacity fits vary scale bounds with budget (Table~\ref{tab:supp_capacity}). The qualitative examples in Fig.~\ref{fig:supp_paintings} use depth-lattice lifting, while quantitative lifting uses perspective back-projection (Fig.~\ref{fig:supp_paintings}, Table~\ref{tab:supp_lift_metrics}). The 2D, lifted-view, and recursive renderers serve fitting, lifting, and mirrors or glass (Fig.~\ref{fig:supp_lift_raytrace}).

Reconstruction scores are per-image fits on observed samples. Image-GS, MiraGe, GaINeR, and \method{} use 500k primitives and 30k steps. GaussianImage uses 70k Gaussians and 50k steps. GaussianImage++ starts at 250k, runs for 30k steps with a 500k cap, and prunes. Instant-NGP uses a $2^{19}$-entry hash table and 30k steps. Kodak-24~\cite{franzen1999kodak} has 24 images at 768$\times$512. DIV2K~\cite{agustsson2017div2k} uses 100 validation images downsampled by two with bicubic~\cite{keys1981cubic} filtering. The default \method{} fit uses a 4D descriptor and source-resolution samples. Capacity and component ablations use a single seed.

\paragraph{Fitting settings and scale bounds}
In the horizontal $[-1,1]$ frame, pixel spacing is $2/W$. The default scale ceiling is about one pixel spacing, with a floor at half that value. We select the best clamped, pre-quantization training fit every 100 steps.

The shared decoder takes a learned four-level 2D grid encoding of local coordinates, a four-frequency ray-direction encoding, and the 4D descriptor. Its two width-64 hidden layers use hard-swish and multiplicatively modulate base color and opacity. The direction stays fixed during fitting and after lifting, making appearance view-independent. Adam~\cite{kingma2015adam} optimizes all parameters with per-group learning rates $\eta_g(t)=\max(\eta_{g,0}\exp(c_g t),\eta_{g,\min})$ over 30k steps, with $c_g=-8.66\times10^{-5}$ except for centers, where $c_g=-1.5\times10^{-4}$. The scale regularizer is $R=\lambda N^{-1}\sum_{i=1}^{N}\|\boldsymbol\sigma_i\|_2$, where $\boldsymbol\sigma_i=\exp(\mathbf s_i)$ is the positive scale vector, $N$ the component count, and $\lambda=5\times10^{-4}$.

\paragraph{Support truncation and compositing}
Each component's ray-tracing proxy is an eight-sided polygon of radius $\sqrt{11.3449}\approx3.37$ in local units, shrunk with opacity so the truncated contribution is below $\exp(-11.3449/2)\approx0.0034$ of its peak opacity weight. Front-to-back compositing stops when transmittance falls below 0.01.

\section{Reconstruction}
This section reports reconstruction variability, baseline protocols, and small-budget fits.
For Table~\ref{tab:reconstruction_seeds}, each method uses its own default seed and the next two. SD is the sample standard deviation of the three dataset means.
\begin{table}[!htb]
\centering\small
\setlength{\tabcolsep}{4pt}
\caption{Reconstruction seeds: mean float PSNR (dB) $\pm$ sample SD over three seeds.}
\label{tab:reconstruction_seeds}
\begin{tabular}{lcc}
\toprule
\textbf{Method} & \textbf{Kodak-24} & \textbf{DIV2K-100}\\
\midrule
SIREN & $37.58\pm0.09$ & $33.42\pm0.06$\\
WIRE & $32.00\pm0.03$ & $29.92\pm0.05$\\
FINER & $39.31\pm0.07$ & $34.26\pm0.04$\\
Instant-NGP & $44.27\pm0.02$ & $41.91\pm0.01$\\
GaussianImage & $44.03\pm0.04$ & $39.52\pm0.01$\\
GaussianImage++ & $53.56\pm0.08$ & $50.48\pm0.10$\\
Image-GS & $50.37\pm0.01$ & $48.58\pm0.00$\\
MiraGe & $71.33\pm0.01$ & $65.66\pm0.01$\\
GaINeR & $72.97\pm0.08$ & $61.34\pm0.01$\\
\method{} & $81.35\pm0.18$ & $70.29\pm0.03$\\
\bottomrule
\end{tabular}
\end{table}

\begin{table*}[!t]
\centering\small
\caption{Kodak-24, default 4D, single seed: best-checkpoint float PSNR per image. The 24-image mean is 81.49\,dB. These values are separate from the three-seed main PSNR.}
\label{tab:exact8_full}
\setlength{\tabcolsep}{6pt}
\begin{tabular}{lr@{\hspace{16pt}}lr}
\toprule
\textbf{Image} & \textbf{PSNR} & \textbf{Image} & \textbf{PSNR}\\
\midrule
kodim01 & 82.62 & kodim13 & 85.69\\
kodim02 & 82.95 & kodim14 & 79.59\\
kodim03 & 79.48 & kodim15 & 76.29\\
kodim04 & 83.68 & kodim16 & 84.09\\
kodim05 & 78.66 & kodim17 & 77.81\\
kodim06 & 83.83 & kodim18 & 78.76\\
kodim07 & 80.58 & kodim19 & 78.16\\
kodim08 & 81.69 & kodim20 & 75.87\\
kodim09 & 84.24 & kodim21 & 83.20\\
kodim10 & 83.60 & kodim22 & 85.87\\
kodim11 & 76.56 & kodim23 & 79.79\\
kodim12 & 88.68 & kodim24 & 84.14\\
\bottomrule
\end{tabular}
\end{table*}
The GaINeR baseline uses the authors' official code at a fixed commit with a 500k budget, 30k steps, no densification or pruning, and one indexing fix in its KNN instance generator. \method{} also exceeds GaINeR's published 500k means: 81.35 versus 77.09\,dB on Kodak and 70.29 versus 62.20\,dB on DIV2K~\cite{jakubowska2026gainer}. Reconstruction PSNR is measured directly during training, before image export.

\paragraph{Additional reconstruction baselines}
GaussianImage~\cite{zhang2024gaussianimage} is reported at its published setting of 70k Gaussians and 50k iterations. Our three-seed Kodak and DIV2K means in Table~\ref{tab:reconstruction_seeds} closely reproduce its published 44.08\,dB on Kodak and 39.53\,dB on DIV2K. At the alternative 500k-Gaussian, 30k-step setting, PSNR is 30.38\,dB on Kodak and 28.63\,dB on DIV2K, close to GaINeR's quoted 30.46\,dB and 28.58\,dB~\cite{jakubowska2026gainer}. The table uses the best observed post-step PSNR, evaluated every 100 iterations. Thus the main comparison does not weaken this baseline by forcing it into the common primitive setting.

GaussianImage++~\cite{li2026gaussianimagepp} uses the authors' official covariance-model code, unmodified apart from a per-image seed for reproducibility, with adaptive densification, pruning, and best-iterate selection. It starts from 250k Gaussians, has a 500k limit, and runs for 30k iterations. Table~\ref{tab:reconstruction_seeds} gives its three-seed quality. Pruning leaves averages of 163k Gaussians and 5.2\,MB on Kodak-24, and 232k Gaussians and 7.4\,MB on DIV2K-100 (7.6\,MB on the DIV2K subset of Table~\ref{tab:reconstruction}). Each Gaussian stores two position, three covariance, and three color float32 values. Training takes 44\,s per Kodak image and 54\,s per DIV2K image on one RTX 4090. The final model is therefore smaller than its initial budget suggests.

MiraGe~\cite{waczynska2025mirage} uses 500k primitives and 30k iterations in 2D-plus-mirror mode, its best-performing mode in our tests. Its three-seed best-training PSNR exceeds both the published GaINeR result and our GaINeR rerun on DIV2K, making it the stronger DIV2K reconstruction comparator. All methods report best training PSNR. Table~\ref{tab:reconstruction_seeds} shows the three-seed variation for \method{}. The descriptor-free control varies more than the full model across three seeds: its means range from 80.14 to 80.72\,dB on Kodak and from 67.85 to 67.99\,dB on DIV2K. Per-image ranges can be larger still (median 1.64\,dB on Kodak). Single-seed ablation differences below about 0.3\,dB on the dataset mean are therefore within seed variation. The descriptor-free ablation remains in the component analysis.

\paragraph{Image-GS reproduction}
We use the official Image-GS~\cite{zhang2025imagegs} code with all 500k Gaussians present from initialization, 30k optimization steps, no progressive addition, no learning-rate decay, and no early stopping. Best training float PSNR is evaluated every 100 iterations. This protocol exceeds the official code's default recipe at a 500k budget (41.73\,dB on Kodak and 42.91\,dB on DIV2K in our runs), so Table~\ref{tab:reconstruction} uses the stronger three-seed result. The Image-GS paper reports no Kodak or DIV2K means for a reproduction check. For DIV2K image \texttt{0843x2} only, we increased the uniform-initialization ratio from 0.30 to 0.73 because the official sampler otherwise fails when too few pixels have nonzero gradient.

\paragraph{Reconstruction comparison}
Table~\ref{tab:exact8_full} lists per-image reconstruction scores from a single seed. The main-table means show that \method{} has the highest PSNR on both datasets. With one seed per method, it exceeds GaINeR, the strongest Kodak baseline, on all 24 Kodak images and MiraGe, the strongest DIV2K baseline, on 97 of 100 DIV2K images. It also exceeds GaINeR on 96 of 100 DIV2K images. This broad per-image pattern is consistent with the shared decoder representing detail within each component while keeping a smaller inference state than the edit-oriented baselines.

\paragraph{Reconstruction error visualization}
The main-text crops in Fig.~\ref{fig:recon-errors} come from Kodak images 5 and 20 at top-left pixel coordinates $(192,64)$ and $(64,128)$. Each map shows the largest rounded eight-bit RGB-channel error, with $2\times2$ max pooling to keep isolated errors visible. Tile labels give the fraction of exact RGB pixels over the full image, whereas the reconstruction table uses training-time float PSNR.

\begin{figure}[!tb]
\centering
\includegraphics[width=\columnwidth,height=.62\textheight,keepaspectratio]{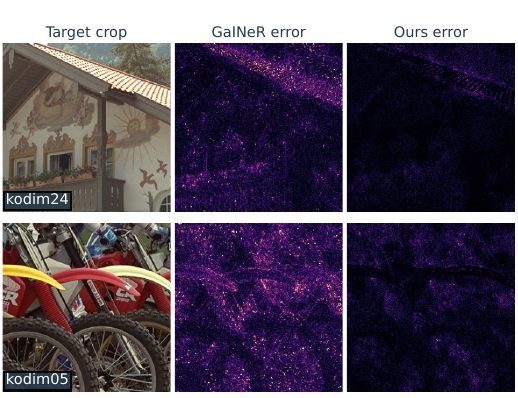}
\caption{Two Kodak-24 crops at 50k primitives (single seed): target and $20\times$ amplified mean absolute RGB error maps for GaINeR and \method{}. All maps use the same scale and colormap. On these images, rounded 8-bit PSNR is 38.8/48.6 dB (image 24) and 39.2/47.5 dB (image 05) for GaINeR/\method{}. The capacity curve reports float PSNR.}
\label{fig:recon50k}

\end{figure}

\paragraph{Coordinate-network INRs with official code}
Table~\ref{tab:reconstruction} reports the same official settings over three seeds on all 100 DIV2K images (Table~\ref{tab:reconstruction_seeds}), while Table~\ref{tab:local_inrs} retains a ten-image subset with its \method{} comparison.
Table~\ref{tab:local_inrs} reports SIREN~\cite{sitzmann2020siren}, WIRE~\cite{saragadam2023wire}, FINER~\cite{liu2024finer}, and Instant-NGP~\cite{mueller2022instant} runs on all 24 Kodak images and DIV2K $2\times$ validation images 0801--0810. We use official model code and optimization settings, with best training float PSNR measured every 100 steps and at the final step, using the same clipped-render metric as \method{}. FINER and its benchmark variants use the official FINER repository. The independent SIREN implementation uses the official SIREN repository. WIRE runs through the FINER benchmark because the WIRE repository has no image-fitting script. Instant-NGP uses the official tiny-cuda-nn~\cite{mueller2021tcnn} image-fitting example. Images are fitted at their native dataset resolution rather than the original papers' 512-pixel crops or resized images, so these results are not directly comparable with their published tables. Their official models are much smaller than \method{}'s inference state. As a memory-matched control, we enlarge only the hash table of the official Instant-NGP configuration from $2^{15}$ to $2^{19}$ entries (30.66\,MB). It improves by 5.6 and 6.5\,dB but remains below \method{} by 37.2 and 29.4\,dB on Kodak-24 and the DIV2K subset. On the full DIV2K-100 validation set it reaches 41.91\,dB.

The \method{} row uses a single seed. Its 71.32\,dB DIV2K mean is restricted to the same ten bicubic-input images and is distinct from the full-validation result. INR training times are CUDA-synchronized per-image means on one RTX 4090 and exclude evaluation renders.

\begin{table*}[!t]
\centering\small
\setlength{\tabcolsep}{4pt}
\caption{Coordinate-network INR fitting with official settings. PSNR (dB) is the mean of best per-image training scores. INR parameter counts and float32 model sizes are shown. The \method{} size is its inference state. Times are seconds per image (Kodak / DIV2K). Official-setting INRs are smaller than \method{}. The enlarged Instant-NGP exceeds its memory. $^\ddagger$Optimized-kernel training (Table~\ref{tab:reconstruction}).}
\label{tab:local_inrs}
\begin{tabular}{p{.32\textwidth}rrll}
\toprule
\textbf{Method (code, steps)} & \textbf{Kodak-24} & \shortstack{\textbf{DIV2K $2\times$}\\\textbf{0801--0810}} & \textbf{Params / MB} & \shortstack{\textbf{Time (s)}\\\textbf{Kodak / DIV2K}}\\
\midrule
SIREN (FINER benchmark, 5k) & 37.10 & 32.97 & 198,915 / 0.80 & 185 / 324\\
SIREN (official SIREN repo, 10k) & 37.48 & 33.20 & 198,915 / 0.80 & 380 / 670\\
WIRE (FINER benchmark, 5k) & 31.99 & 29.26 & 199,287 / 0.80 & 640 / 1139\\
FINER (official FINER repo, 5k) & 39.25 & 34.16 & 198,915 / 0.80 & 253 / 447\\
Instant-NGP (official tcnn, 30k) & 38.69 & 35.42 & 715,536 / 2.86 & 28 / 28\\
Instant-NGP (tcnn, $2^{19}$ hash table, 30k) & 44.26 & 41.88 & 7,665,552 / 30.66 & 37 / 38\\
\midrule
\method{} 4D (single seed) & \best{81.49} & \best{71.32} & 7.02M / 28.08 & 929 / 1218$^\ddagger$\\
\bottomrule
\end{tabular}
\end{table*}

\paragraph{Small-budget reconstruction}
Fig.~\ref{fig:recon50k} visualizes the 50k capacity comparison. Their amplified error maps share the same scale. The caption identifies the rounded-8-bit metric, distinct from the float PSNR in the capacity curve.

\section{Appearance and transport}
We first examine the component budget, then test frame transport and smudge editing with their controls.
\subsection{Component analysis and budget settings}
Table~\ref{tab:supp_ablations} retains all component controls from the single-seed Kodak ablation. Constant per-primitive color keeps learned base RGB, opacity, and spatial support while disabling neural modulation, leaving the descriptor inactive. Other rows disable the named component. Removing local appearance or local coordinates causes the largest losses, consistent with detail being decoded inside each component. The single-seed descriptor gap differs from the three-seed gains of 0.98\,dB on Kodak and 2.37\,dB on DIV2K. Smaller ablation differences should be read against the seed variation above. Best comparable results are bold, and near-ties remain unranked.
\begin{table}[t]
\centering\small
\setlength{\tabcolsep}{4pt}
\caption{Complete component analysis on Kodak-24, 500k primitives, a single seed. PSNR and changes from the full model are in dB. Constant per-primitive color disables neural color and opacity modulation; the descriptor is inactive. Each other row disables the named component.}
\label{tab:supp_ablations}
\begin{tabular}{lrr}
\toprule
\textbf{Variant} & \textbf{PSNR} & \textbf{$\Delta$}\\
\midrule
\textbf{Full FACET} & \best{81.49} & --\\
Constant per-primitive color & 72.19 & -9.30\\
Without local $uv$ & 73.64 & -7.85\\
Without descriptor & 80.14 & -1.35\\
Without scale floor & 80.60 & -0.89\\
Without learned angle & 80.98 & -0.52\\
Without opacity modulation & 81.13 & -0.36\\
Without generalized shape & 81.45 & -0.05\\
Without scale regularization & 81.43 & -0.07\\
\bottomrule
\end{tabular}
\end{table}

\paragraph{Per-component texture at matched memory}
Table~\ref{tab:component_texture} compares single-seed best training PSNR on Kodak-24. The texture replaces the shared decoder with a bilinearly sampled $4\times4$ RGB map per component and uses the \method{} recipe. Its 127,649 components match the full model's 28.08\,MB inference memory. Fewer components make portrait coverage sensitive to scale bounds, so we measure the texture's cap on the long image side. At most 0.7\% of full-model components reach the cap. At equal memory, sharing the decoder across components leaves room for more components and a more accurate fit.

\begin{table}[!htb]
\centering\small
\setlength{\tabcolsep}{4pt}
\caption{Per-component appearance comparison on Kodak-24, single seed. PSNR is in dB. The texture and shared-decoder models have matched inference memory.}
\label{tab:component_texture}
\begin{tabular}{lrr}
\toprule
\textbf{Appearance (components)} & \textbf{Mean} & \textbf{Median}\\
\midrule
\method{} 4D (500k) & \best{81.49} & \best{82.15}\\
$4\times4$ RGB texture (127,649) & 78.13 & 78.34\\
Constant color (500k) & 72.19 & 69.56\\
\bottomrule
\end{tabular}
\end{table}

Table~\ref{tab:supp_capacity} retains the full model, no-descriptor ablation, and GaINeR across budgets. At 50k components the no-descriptor variant falls below GaINeR, while the full model exceeds it, supporting the role of learned local appearance at a small budget. We adjust the \method{} scale cap as $0.0025\sqrt{500{,}000/N}$, with a floor at half the cap, so this is a budget-dependent recipe rather than a count-only sweep. Across all tested budgets, \method{} exceeds GaINeR on every image with less memory. Its shared decoder also leads the per-component texture by 3.4\,dB at matched memory, while MiraGe densifies to a much larger state.
\begin{table}[t]
\centering\small
\caption{Kodak-24 capacity results (single seed). All scores are mean float PSNR (dB). The no-descriptor variant is retained here as an ablation.}
\label{tab:supp_capacity}
\begin{tabular}{rrrr}
\toprule
\textbf{Primitives} & \textbf{FACET} & \textbf{No descriptor} & \textbf{GaINeR}\\
\midrule
50k & \best{49.82} & 39.73 & 42.51\\
100k & \best{61.22} & 52.31 & 49.76\\
250k & \best{77.60} & 74.13 & 62.85\\
500k & \best{81.49} & 80.14 & 73.03\\
\bottomrule
\end{tabular}
\end{table}

\subsection{Controlled transport}
The controlled transport evaluation uses single-seed fits of Kodak images 01, 05, and 23. Rotation is 30 degrees, horizontal stretch is 2.5$\times$, and the swirl rotates by 90 degrees within a 240-pixel radius. Fig.~\ref{fig:transport} uses a stronger $180^\circ$ swirl within a 160-pixel radius from the same fits. Each target is the original eight-bit image sampled bicubically through the inverse warp. PSNR includes only pixels mapped at least two source pixels inside the image, so it reflects both fitting error and target interpolation. Table~\ref{tab:transport_full} shows that carrying each component's frame consistently beats moving only its center, especially where stretch would otherwise open gaps between supports.

\begin{table}[!htb]
\centering\small
\caption{Transport versus moving centers only on Kodak-24 images 01, 05, and 23, default 4D single-seed fit. PSNR is computed on pixels with valid inverse-mapped source locations.}
\label{tab:transport_full}
\setlength{\tabcolsep}{4pt}
\begin{tabular}{llrrr}
\toprule
\textbf{Image} & \textbf{Warp} & \textbf{Centers} & \textbf{Transport} & \textbf{Gain}\\
\midrule
kodim01 & rotate & 34.27 & \best{37.70} & +3.44\\
kodim01 & scale & 13.49 & \best{39.90} & +26.41\\
kodim01 & swirl $90^\circ$ & 33.45 & \best{40.79} & +7.35\\
kodim01 & swirl $180^\circ$ & 32.15 & \best{43.71} & +11.56\\
kodim05 & rotate & 34.27 & \best{38.01} & +3.75\\
kodim05 & scale & 14.62 & \best{39.37} & +24.75\\
kodim05 & swirl $90^\circ$ & 33.22 & \best{40.61} & +7.39\\
kodim05 & swirl $180^\circ$ & 31.44 & \best{43.70} & +12.26\\
kodim23 & rotate & 40.33 & \best{42.18} & +1.85\\
kodim23 & scale & 12.56 & \best{43.45} & +30.88\\
kodim23 & swirl $90^\circ$ & 37.94 & \best{44.81} & +6.87\\
kodim23 & swirl $180^\circ$ & 37.94 & \best{48.44} & +10.50\\
\bottomrule
\end{tabular}
\end{table}

\paragraph{Transport components and shear}
Table~\ref{tab:transport_components} uses the same fits, reference, and valid-pixel rule. Every variant moves centers. Rotation-only transport updates the angle, scales-only transport updates the scales, and full transport updates both. We also test horizontal shear $(x,y)\mapsto(x+ky,y)$ about the image center. Rotation alone suffices for pure rotation, and scales alone recover some coverage under stretch. Both updates are needed under stretch, swirl, and shear, consistent with a decoded pattern that must move with the whole local frame.

Let $\mathbf A=\mathbf F\mathbf R(\theta)\operatorname{diag}(\mathbf s)$ be the exact local affine map and $\mathbf A'=\mathbf R(\theta')\operatorname{diag}(\mathbf s')$ its representable frame, where $\mathbf s,\mathbf s'$ denote positive scales. Both map the unit disk to the same ellipse, but the residual rotation $\mathbf Q=\mathbf A'^{-1}\mathbf A$ changes the local coordinates queried by the decoder. We report the absolute angle of $\mathbf Q$: this measures the local-coordinate rotation omitted by the SVD frame approximation under any non-similarity map. It averages about $22^\circ$ under stretch and shear. Full transport reaches 39.22--41.55\,dB on the shear sweep, and PSNR decreases by 2.32\,dB as shear grows from $k=0.25$ to $k=1.0$. This decrease reflects the harder warp. The exact-affine query below isolates the frame approximation and changes PSNR by at most 0.14\,dB.

\begin{table*}[!t]
\centering\small
\setlength{\tabcolsep}{6pt}
\caption{Transport controls on three Kodak images. The first four columns report mean PSNR (dB). The last two report residual frame angles in degrees, averaged over the per-image means and per-image 95th percentiles. All variants use the same checkpoints and references.}
\label{tab:transport_components}
\begin{tabular}{lrrrrrr}
\toprule
\textbf{Warp} & \textbf{Centers only} & \textbf{Rotation only} & \textbf{Scales only} & \textbf{Full transport} & \textbf{Mean angle} & \textbf{p95 angle}\\
\midrule
Rotation $30^\circ$ & 36.29 & 39.30 & 36.29 & 39.30 & 0.01 & 0.00\\
Stretch $2.5\times$ & 13.56 & 13.57 & 28.65 & 40.91 & 22.07 & 42.95\\
Swirl $90^\circ$ & 34.87 & 35.66 & 34.83 & 42.07 & 9.78 & 39.53\\
Shear $k=0.25$ & 38.21 & 38.71 & 38.25 & 41.55 & 20.58 & 42.23\\
Shear $k=0.5$ & 33.85 & 34.63 & 33.49 & 41.26 & 21.98 & 42.53\\
Shear $k=1.0$ & 28.77 & 29.25 & 26.98 & 39.22 & 22.44 & 43.02\\
\bottomrule
\end{tabular}
\end{table*}

\paragraph{Full affine decoder query}
Querying the decoder through the exact local affine map instead of the SVD frame removes the residual rotation while keeping the same ellipse support. On the warp tests above it improves PSNR by at most 0.14\,dB, at about 1\% additional render time, and changes brushed-region PSNR in the smudge benchmark by $-0.02$\,dB. We therefore keep the SVD frame.

\paragraph{Smudge-brush protocol}
Scoring each edit against the inverse warp of its method's undeformed render changes brushed-region PSNR by at most 0.06\,dB and leaves all rankings unchanged. On frozen geometry, a better-fitting texture still loses every case (Table~\ref{tab:texture_geometry}).
The benchmark uses landscape Kodak images 03, 05, 07, 14, 20, and 21 and one FACET and GaINeR fit per image. Two six-stroke sets are applied to every image. Each stroke is a Catmull--Rom curve through four random points with brush radius $R$ between 28 and 48 pixels. We advance in steps of $R/4$ and apply radial displacement $s(1-(r/R)^2)^2$ inside the brush, with $s=0.85$. Composing the inverse displacements gives the bicubic-warped input reference. PSNR is measured over valid pixels and over those moved by more than 0.5 pixels. Table~\ref{tab:smudge} averages the 12 image--stroke-set pairs. The 2D Gaussian baselines use single-seed reconstruction fits: their centers follow the warp and covariances transform as $\mathbf J\boldsymbol\Sigma\mathbf J^\top$ at each center. Moving only their centers yields 12.02\,dB for GaussianImage, 12.86\,dB for GaussianImage++, and 23.66\,dB for Image-GS on brushed pixels. \method{} wins every case against each transported baseline. Fig.~\ref{fig:smudge} shows one measured case and two showcases under this protocol.

\paragraph{Six-image results}
Table~\ref{tab:smudge} shows transport winning every case against center-only motion, GaINeR, and the matched-memory $4\times4$ texture, which uses the same Jacobian transport and warps. The full affine query changes brushed-region PSNR by only $-0.02$\,dB. On Kodak 20, the texture fits better (77.5 versus 75.9\,dB) yet loses after the two strokes by 5.3 and 2.3\,dB. The editing gap therefore does not simply inherit reconstruction error. This supports frame-attached appearance as the source of the editing advantage.

\begin{table}[!htb]
\centering\small
\setlength{\tabcolsep}{3.8pt}
\caption{Smudge-brush PSNR (dB) on six Kodak images and two six-stroke sets. Image: valid pixels. Brushed: valid pixels moved by more than 0.5 pixels. $^*$70k Gaussians, 50k steps (its reconstruction-table setting).}
\label{tab:smudge}
\begin{tabular}{lrr}
\toprule
\textbf{Variant} & \textbf{Image} & \textbf{Brushed}\\
\midrule
Centers only & 22.91 & 15.63\\
Transport & \best{39.54} & \best{32.31}\\
Full affine & 39.53 & 32.29\\
GaINeR & 29.70 & 22.50\\
GaussianImage$^*$ & 22.86 & 16.23\\
GaussianImage++ & 21.02 & 14.47\\
Image-GS & 34.68 & 28.29\\
$4\times4$ texture (matched memory) & 36.63 & 29.42\\
\midrule
\multicolumn{3}{l}{\textbf{Smaller budgets (image / brushed)}}\\
\method{} transport, 50k & \best{31.11} & \best{24.05}\\
GaINeR, 50k & 17.07 & 11.13\\
\method{} transport, 100k & \best{34.27} & \best{27.11}\\
GaINeR, 100k & 21.71 & 14.52\\
\bottomrule
\end{tabular}
\end{table}

\paragraph{Wider benchmark and seeds}
We extend the same protocol to all 24 Kodak images and four six-stroke sets, giving 96 cases. Portraits use a $512\times768$ canvas and the texture's long-side scale cap from Table~\ref{tab:component_texture}. Transported \method{} reaches 32.59\,dB on brushed pixels, above both GaINeR (22.51\,dB) and the texture (30.32\,dB). It wins all 96 cases against GaINeR and 91 against the texture, including every landscape case. The texture wins five portrait cases by 0.12--1.44\,dB on brushed pixels. Two additional \method{} seeds reach 32.41 and 32.40\,dB and also beat the single-seed GaINeR result in every case, by at least 2.33\,dB. Their per-case range is 0.47\,dB on average and at most 3.17\,dB. Against the texture, the three \method{} seeds win 91, 88, and 86 of 96 cases. The stable advantage over GaINeR is consistent with appearance being decoded in frames carried by the moving components, while the texture's few portrait wins show that the margin is not universal.

\paragraph{Texture on the same geometry}
Table~\ref{tab:texture_geometry} separates appearance from geometry on six landscape images and four stroke sets. The same-geometry rows copy and freeze geometry and base opacity from the \method{} 4D fit. We refit only appearance for 30k iterations: the shared decoder, a $4\times4$ RGB texture, or the decoder with opacity output fixed so that it changes only color. The texture fits better but loses after all 24 smudges, including against the color-only decoder. A more accurate source fit does not by itself preserve detail under deformation.

\begin{table}[t]
\centering\footnotesize
\setlength{\tabcolsep}{3pt}
\caption{Texture versus shared decoder on six Kodak images (03, 05, 07, 14, 20, 21), single seed. Fit is best training PSNR. Smudge is brushed-region PSNR over 24 cases (four stroke sets per image), against the exact inverse warp of the input. Wins counts the cases in which a row beats the texture of its block.}
\label{tab:texture_geometry}
\begin{tabular}{lrrccc}
\toprule
\textbf{Appearance} & \textbf{Comp.} & \textbf{MB} & \textbf{Fit} & \textbf{Smudge} & \textbf{Wins}\\
\midrule
\multicolumn{6}{l}{\emph{Matched memory, own geometry}}\\
$4\times4$ texture & 127,649 & 28.1 & 77.25 & 30.51 & --\\
\method{} & 500,000 & 28.1 & \best{79.56} & \best{33.26} & 24/24\\
\midrule
\multicolumn{6}{l}{\emph{Same frozen geometry}}\\
$4\times4$ texture & 500,000 & 110.0 & \best{80.17} & 32.60 & --\\
Shared decoder & 500,000 & 28.1 & 76.75 & 33.55 & 24/24\\
Shared decoder, color only & 500,000 & 28.1 & 76.31 & \best{33.59} & 24/24\\
\bottomrule
\end{tabular}
\end{table}

\paragraph{Smaller budgets}
At 50k and 100k components we repeat the 12-case benchmark using \method{} capacity-curve fits and GaINeR retrained at equal budgets with a single seed. Transport wins every case at both budgets, and its brushed-region margin over GaINeR increases as the budget shrinks. At matched inference memory (2.9 and 5.7\,MB), the $4\times4$ texture also loses every case. It fits the source images 11.9 and 14.3\,dB worse at the two budgets. Each texture uses the better of two scale-bound settings. Sharing the decoder remains useful when few components and little memory are available.

\section{Lifting}
We report held-out views, controls, source ordering, and ray queries. We share GaINeR's single-image lifting task and Depth Pro source~\cite{jakubowska2026gainer}, but define our own views, GT-depth comparison, and metrics below.
The quantitative Hotdog and Lego (NeRF-Synthetic~\cite{mildenhall2020nerf}) and DTU~\cite{jensen2014dtu} scan114 and scan110 results use one fitted source view per scene and both ground-truth and Depth Pro~\cite{bochkovskii2025depthpro} source depth. Hotdog uses source r\_14 and targets r\_32--r\_39. Lego uses source r\_90 and targets r\_23--r\_30. Each DTU scene uses source view 25 and targets 24 and 26. Table~\ref{tab:lift} reports PSNR over full output images. Components with invalid depth are deactivated. At steep depth edges, a minmod slope limiter bounds the depth gradient at an $85^\circ$ slant threshold without removing components.

\paragraph{GT-depth completion inside the mask}
For all four quantitative scenes, missing GT depth inside the source-view object mask is filled from the nearest valid depth, identically for both methods. Background pixels stay invalid. The filled-pixel counts are 414 for Hotdog, 953 for Lego, 46 for DTU 114, and 8307 for DTU 110 (4.7\% of its source-object mask). Co-visibility masks are computed from the original depth. Depth Pro rows are unaffected. The remaining GT-depth holes on DTU 110 lie at steep depth edges, which Depth Pro covers more completely.

\paragraph{Full lifting metrics}
Table~\ref{tab:supp_lift_metrics} reports full-image and source-visible PSNR, full-image SSIM, and full-image LPIPS for eight scene--depth settings. A target pixel is source-visible when a source-object pixel with valid GT depth projects into it after a $3\times3$ morphological closing. Across 40 held-out views at 10--18$^\circ$ from the source camera, \method{} exceeds GaINeR in 39 and the mesh in 39. Its margin over GaINeR narrows from 12.1\,dB at $10^\circ$ to 7.9\,dB at $18^\circ$ on Hotdog with GT depth, consistent with an increasing contribution from unseen geometry that neither single-view fit can recover.

\begin{table*}[!t]

\centering\small
\setlength{\tabcolsep}{5pt}
\caption{Lifting metrics with in-mask filled GT depth (GT) or Depth Pro (DP), $\delta=0.01$. Each cell is GaINeR / FACET. Covis: pixels visible from the source view. Higher is better except LPIPS.}
\label{tab:supp_lift_metrics}
\begin{tabular}{llrrrr}
\toprule
\textbf{Scene} & \textbf{Depth} & \textbf{PSNR full} & \textbf{PSNR covis} & \textbf{SSIM} & \textbf{LPIPS}\\
\midrule
Hotdog & GT & 20.15 / \best{30.65} & 20.44 / \best{29.97} & .844 / \best{.962} & .183 / \best{.061}\\
Hotdog & DP & 19.29 / \best{20.68} & 19.87 / \best{20.64} & .839 / \best{.876} & .204 / \best{.141}\\
Lego & GT & 16.27 / \best{21.41} & 16.03 / \best{23.79} & .684 / \best{.891} & .261 / \best{.120}\\
Lego & DP & 15.65 / \best{15.78} & \best{12.90} / 12.01 & .685 / \best{.710} & .258 / \best{.200}\\
DTU 114 & GT & 17.31 / \best{20.55} & 19.24 / \best{25.66} & .629 / \best{.883} & .310 / \best{.128}\\
DTU 114 & DP & 16.48 / \best{19.04} & 18.49 / \best{20.75} & .603 / \best{.738} & .327 / \best{.175}\\
DTU 110 & GT & 13.91 / \best{16.74} & 17.47 / \best{21.11} & .751 / \best{.854} & .293 / \best{.158}\\
DTU 110 & DP & 13.43 / \best{19.16} & 17.37 / \best{19.54} & .743 / \best{.837} & .296 / \best{.150}\\
\bottomrule
\end{tabular}
\end{table*}

\subsection{Additional lifting views}
\begin{figure*}[!htb]
\centering
\includegraphics[width=.94\textwidth,height=.62\textheight,keepaspectratio]{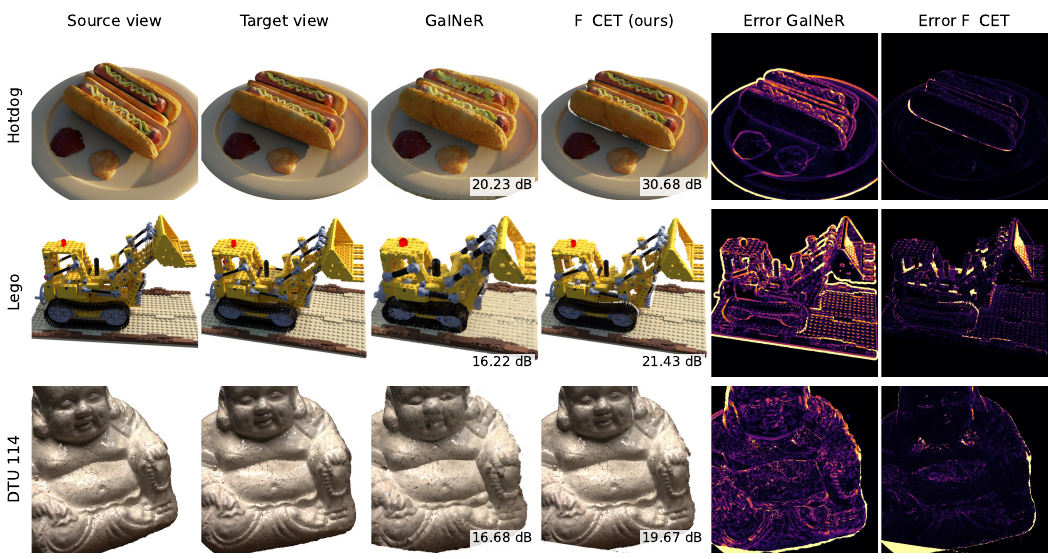}
\caption{Single-image lifting on Hotdog, Lego, and DTU 114 with error maps on a shared scale. Panels use the same filled GT-depth renders as Table~\ref{tab:lift}. Tile scores are per-view PSNR.}
\label{fig:supp_lift_full}

\end{figure*}

This subsection collects qualitative lifting examples under the protocols above.
Fig.~\ref{fig:supp_lift_full} shows held-out views of Hotdog, Lego, and DTU 114. Fig.~\ref{fig:lift} shows Hotdog and DTU 114 from this comparison.

Hotdog and Lego use the lower-median \method{}-PSNR view among eight targets, while DTU 114 uses its first target view. The selected views are Hotdog \texttt{r\_34}, Lego \texttt{r\_28}, and DTU 114 view 24. Center-crop zoom factors are 1.35, 1.25, and 1.0, respectively. Cropping does not change the full-image scores, and Table~\ref{tab:lift} retains all four scenes and both depth sources.

\paragraph{Texture versus \method{} after lifting}
Table~\ref{tab:supp_texture_lift} compares \method{} with the matched-memory $4\times4$ texture of Table~\ref{tab:component_texture}, using the same renderer, index offset, random-background fit, and depth protocol. \method{} leads on every metric in all eight settings. On identical frozen geometry with appearance refitted, the texture has 0.02--0.12\,dB higher full-image PSNR in all eight settings, while the decoder has lower LPIPS in seven. The texture uses 110\,MB for the same 500k components. This suggests the matched-memory advantage comes chiefly from fitting more geometry and coverage with the shared decoder, rather than a PSNR advantage on identical geometry.

The texture model's scale cap and floor are scaled to preserve the per-component pixel coverage of the Kodak texture baseline: $1.28\times$ for the $800\times800$ synthetic images. DTU has the Kodak resolution and uses the Kodak bounds.

\begin{table*}[!t]
\centering\small
\setlength{\tabcolsep}{5pt}
\caption{Texture versus FACET after lifting. Each cell is FACET / texture. Covis: pixels visible from the source view. Higher is better except LPIPS.}
\label{tab:supp_texture_lift}
\begin{tabular}{llrrrr}
\toprule
\textbf{Scene} & \textbf{Depth} & \textbf{PSNR full} & \textbf{PSNR covis} & \textbf{SSIM} & \textbf{LPIPS}\\
\midrule
Hotdog & GT & \best{30.65} / 26.21 & \best{29.97} / 26.28 & \best{.962} / .899 & \best{.061} / .159\\
Hotdog & DP & \best{20.68} / 19.89 & \best{20.64} / 19.20 & \best{.876} / .816 & \best{.141} / .219\\
Lego & GT & \best{21.41} / 19.44 & \best{23.79} / 20.33 & \best{.891} / .845 & \best{.120} / .172\\
Lego & DP & \best{15.78} / 15.48 & \best{12.01} / 11.74 & \best{.710} / .693 & \best{.200} / .231\\
DTU 114 & GT & \best{20.55} / 19.21 & \best{25.66} / 21.81 & \best{.883} / .755 & \best{.128} / .261\\
DTU 114 & DP & \best{19.04} / 18.24 & \best{20.75} / 19.07 & \best{.738} / .643 & \best{.175} / .294\\
DTU 110 & GT & \best{16.74} / 15.69 & \best{21.11} / 18.47 & \best{.854} / .700 & \best{.158} / .271\\
DTU 110 & DP & \best{19.16} / 18.37 & \best{19.54} / 17.53 & \best{.837} / .683 & \best{.150} / .273\\
\bottomrule
\end{tabular}
\end{table*}

\begin{figure*}[!htb]
\centering
\includegraphics[width=.95\textwidth,height=.62\textheight,keepaspectratio]{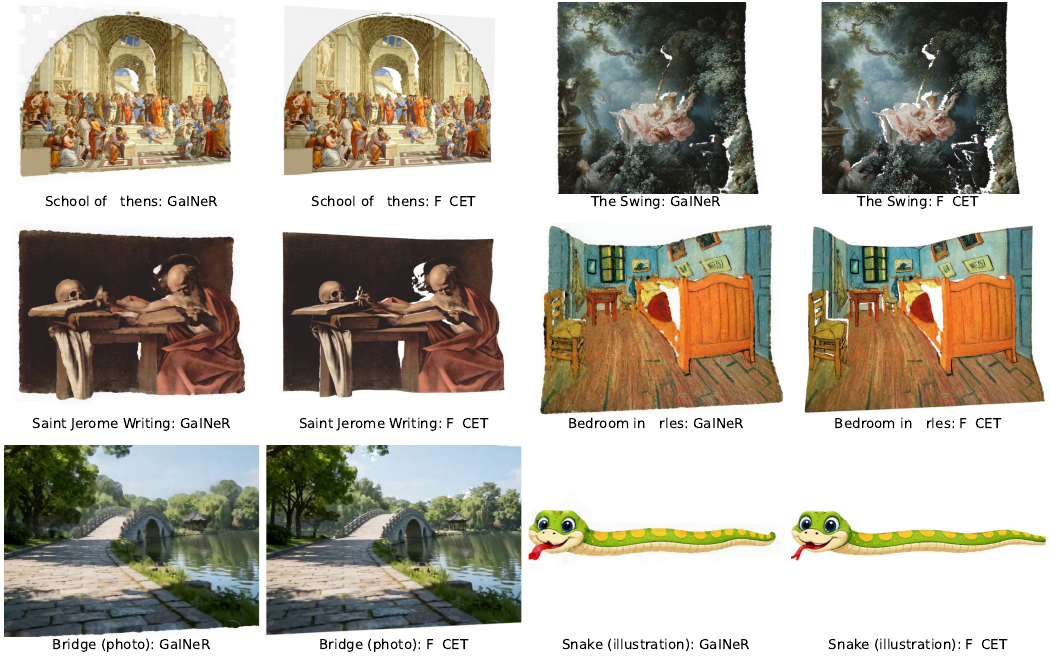}
\caption{Qualitative lifting of paintings and photographs, GaINeR versus FACET at the same depth and camera pose (depth-lattice protocol in Sec.~\ref{sec:depth_lattice}).}
\label{fig:supp_paintings}
\end{figure*}

\paragraph{Qualitative depth-lattice lifting}\label{sec:depth_lattice}
Fig.~\ref{fig:supp_paintings} compares lifted paintings and photographs at matched depth and camera pose. These demonstrations use a depth-lattice lift matched to GaINeR's deformation conventions, rather than the perspective back-projection of the quantitative experiments.

The supplied depth displaces the image lattice. This leaves the appearance fit unchanged, so differences in the gallery reflect how the two representations respond to the same displaced lattice.

\paragraph{Textured depth mesh}
Table~\ref{tab:supp_mesh_lift} compares \method{} with a textured triangle mesh built from the same source image and filled GT or Depth Pro depth. Camera intrinsics, back-projection, target cameras, and evaluation match the protocol of Table~\ref{tab:lift}. Each source-pixel quad gives two triangles, cast with Open3D~\cite{zhou2018open3d}. The mesh drops triangles with invalid depth or normals beyond the $85^\circ$ threshold, while \method{} limits the depth gradient at that threshold with a minmod slope limiter. Source RGB and alpha are interpolated barycentrically. Sampling source RGB bilinearly as a UV texture instead changes full-image PSNR by at most 0.01\,dB in all eight settings. Alpha comes from RGBA sources and is one for opaque photographs. There is no texture prefiltering, and the first hit is composited over the background.

\method{} has higher full-image and source-visible PSNR in every setting and fewer holes. The mesh has higher SSIM on all four DTU settings and lower LPIPS in three settings. Its sharp source-visible texture can favor structural metrics, but slant cuts and hidden surfaces leave uncovered target pixels. Overlapping components can cover some of these gaps with partial opacity. Thus \method{} keeps higher pixel accuracy and coverage, while the mesh remains competitive on structural metrics for real photographs.

\begin{table*}[!t]
\centering\small
\setlength{\tabcolsep}{3pt}
\caption{Lifting comparison with a textured depth mesh built from the same source image and depth. Each cell is FACET / mesh. Covis: pixels visible from the source view. Holes: target pixels inside the object left uncovered. Higher is better except LPIPS and holes.}
\label{tab:supp_mesh_lift}
\begin{tabular}{llrrrrr}
\toprule
\textbf{Scene} & \textbf{Depth} & \textbf{PSNR full} & \textbf{PSNR covis} & \textbf{SSIM} & \textbf{LPIPS} & \textbf{Holes (\%)}\\
\midrule
Hotdog & GT & \best{30.65} / 22.95 & \best{29.97} / 22.79 & \best{.962} / .956 & \best{.061} / .073 & 0.1 / 2.3\\
Hotdog & DP & \best{20.68} / 18.85 & \best{20.64} / 17.68 & \best{.876} / .875 & .141 / \best{.134} & 0.7 / 3.7\\
Lego & GT & \best{21.41} / 16.45 & \best{23.79} / 14.82 & \best{.891} / .859 & \best{.120} / .138 & 2.8 / 13.9\\
Lego & DP & \best{15.78} / 14.66 & \best{12.01} / 11.05 & \best{.710} / .707 & \best{.200} / .202 & 8.6 / 15.8\\
DTU 114 & GT & \best{20.55} / 19.16 & \best{25.66} / 22.91 & .883 / \best{.888} & \best{.128} / .133 & 1.6 / 2.9\\
DTU 114 & DP & \best{19.04} / 18.55 & \best{20.75} / 19.96 & .738 / \best{.754} & .175 / \best{.166} & 1.7 / 3.1\\
DTU 110 & GT & \best{16.74} / 15.14 & \best{21.11} / 17.94 & .854 / \best{.855} & \best{.158} / .183 & 2.3 / 5.7\\
DTU 110 & DP & \best{19.16} / 18.45 & \best{19.54} / 17.97 & .837 / \best{.859} & .150 / \best{.124} & 0.6 / 2.3\\
\bottomrule
\end{tabular}
\end{table*}

\paragraph{Sensitivity to the index-order offset}
The lifted renderer sorts hits by $t+\delta\,\rho\,\mathrm{id}/2^{19}$, where $t$ is the ray-hit distance, $\rho$ the hit's distance from the source camera, and $\mathrm{id}$ the component index. We use $\delta=0.01$, the smallest value tested. Since component indices stay below $2^{19}$, it perturbs hit distances by less than 1\% of $\rho$. Consequently, two hits can swap order only when their distances along the ray differ by less than 1\% of $\rho$. Surfaces separated by more than that keep their geometric order, and the offset only decides the order of nearly coincident supports, which is where the source-view order matters. In a controlled test, two fitted Kodak images (250k components each) are lifted onto parallel planes at depths 1 and $1+g$ and rendered from the source camera and from a camera rotated by $10^\circ$, against a reference that composites the two layers in exact depth order. When the component indices favor the front layer, every gap from $g=0.5$ down to $0.001$ matches the reference (87--89\,dB, no changed pixels). When they favor the back layer, every gap down to $g=0.01$, 1\% of the distance, still matches, and the layers swap at $g=0.005$. Because the offset is scaled by the distance from the source camera rather than from the ray origin, reflected and refracted rays use the same tolerance as camera rays. Moving all ray origins onto a mirror plane at distance $m=0.5$, $0.2$, $0.1$, or $0.05$ in front of the layers confirms this. With the back layer favored, every origin matches the reference down to $g=0.01$ and swaps at $g=0.005$, exactly as the source camera does. With the front layer favored, no gap swaps for any origin.
Table~\ref{tab:id_order_offset} varies this offset on the four GT-depth scenes with in-mask filled depth and the same evaluation as the main lifting table. Across the $30\times$ range $\delta=0.01$--$0.3$, full-image PSNR spans 0.39\,dB on Hotdog, 0.29\,dB on Lego, and at most 0.11\,dB on the DTU scenes. Larger offsets in this range are slightly better than the default, which we keep as the smallest tested value rather than tuning it on the evaluation views. Removing the offset lowers full-image PSNR by 1.2--2.6\,dB and SSIM by 0.08--0.23, and raises LPIPS by 0.08--0.23 relative to the default. To test temporal stability, we render each lifted scene along a continuous orbit (yaw $-12^\circ$ to $12^\circ$ in $0.2^\circ$ steps) and count popping pixels, whose second temporal difference exceeds 0.1 in any channel. With $\delta=0.01$, 0.9--2.7\% of covered pixels pop per frame, versus 1.9--6.3\% without the offset. Halving the step reduces the remaining pops 3.2--8.7$\times$ with the offset versus 1.9--4.2$\times$ without it, as expected if fewer of them come from order changes. The offset therefore supports source-view ordering, while performance remains stable over a broad range of nonzero values.

\begin{table}[!htb]
\centering\small
\setlength{\tabcolsep}{3pt}
\caption{Full-image PSNR (dB) for the index-order offset with filled GT depth. The default is $\delta=0.01$. ``Off'' sorts by hit distance with an index tie break.}
\label{tab:id_order_offset}
\begin{tabular}{lrrrrr}
\toprule
\textbf{Scene} & \textbf{Off} & $\boldsymbol\delta=0.01$ & $\boldsymbol\delta=0.03$ & $\boldsymbol\delta=0.1$ & $\boldsymbol\delta=0.3$\\
\midrule
Hotdog & 28.01 & 30.65 & 30.81 & 30.91 & 30.52\\
Lego & 20.18 & 21.41 & 21.61 & 21.70 & 21.69\\
DTU 114 & 18.27 & 20.55 & 20.63 & 20.66 & 20.66\\
DTU 110 & 15.43 & 16.74 & 16.75 & 16.76 & 16.76\\
\bottomrule
\end{tabular}
\end{table}

\subsection{Ray queries}
\begin{figure*}[!htb]
\centering
\includegraphics[width=\textwidth,height=.62\textheight,keepaspectratio]{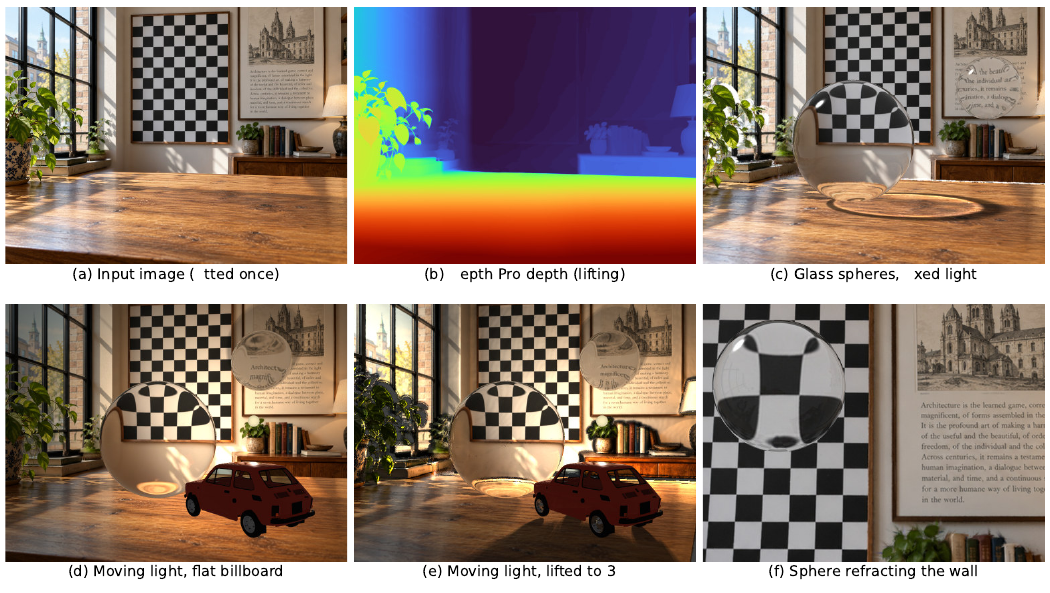}
\caption{\textbf{Single-image lifting and ray tracing.} (a) Input image after SDXL chair removal, before fitting. (b) Depth Pro depth used for lifting. (c) Inserted glass spheres under a fixed light. (d,e) The same moving light applied to the image as a flat billboard and to the lifted field: the scene's depth-dependent shadows and occlusions appear after lifting. (f) A sphere refracts the wall. The fitted appearance is retained during geometric lifting and ray tracing. The panels are qualitative outputs of the recursive ray tracer.}
\label{fig:supp_lift_raytrace}
\end{figure*}

Fig.~\ref{fig:supp_lift_raytrace} follows the fitted checkerboard image from its input and supplied depth to a ray-traced scene. The chair is removed with SDXL inpainting~\cite{podell2024sdxl,diffusers2023sdxlinpaint} before fitting. This preprocessing is not performed by \method{}. Depth Pro~\cite{bochkovskii2025depthpro} then supplies the geometry for lifting. With appearance held fixed, the lifted field supports intersections with rays from inserted glass and casts geometry-dependent shadows. The depth panel visualizes log depth with a common percentile normalization (near surfaces are warm). The flat-billboard and lifted panels use the same moving light: lifting introduces the scene's depth-dependent shadows and occlusions. This is a qualitative demonstration rendered with the recursive ray tracer.

\section{Cost}
This section records training, memory, and rendering costs for the comparisons in the main text.
Training time and memory are per-image means on Kodak-24. The coordinate-network and 2D Gaussian render rates are measured with each method's own renderer right after fitting, using the same timing protocol.
DIV2K costs use validation images 0801--0810 for training time and memory and images 0801, 0807, and 0809 for render rates, with the same protocol as Kodak. MiraGe's DIV2K memory is larger than on Kodak because densification adds more Gaussians on the larger images (258.9\,MB on average).

Memory in the main cost table is the uncompressed inference state rather than the optimizer checkpoint. MiraGe densifies from 500k initialized Gaussians to 0.86M on average on Kodak-24. Each Gaussian stores 62 float32 values (position, scale, rotation, opacity, and degree-3 SH color). In particular, the 28 MB per-image state is designed for editing and ray tracing, not image compression. We do not evaluate a quantized or entropy-coded bitstream. The three-seed reconstruction table uses the reference kernel. Optimized-kernel quality checks and rendering rates are specified below.

\paragraph{Kernel implementation and rendering cost}
The optimized 4D kernel uses less peak training memory than GaINeR on one RTX 4090: 906 versus 3969\,MiB on Kodak image 01 and 1423 versus 6436\,MiB on DIV2K image 0801. Its single-seed Kodak quality is 81.38\,dB, within the range of the reference-kernel seed means. The reconstruction table retains reference-kernel scores. Both kernels use the same model state, although floating-point evaluation order differs. Static render rates in Table~\ref{tab:reconstruction} are medians of 20 warmed CUDA-timed renders, averaged over Kodak images 05, 18, and 19. GaINeR uses a cached KNN state, MiraGe uses rasterization, and \method{} uses its reference 2D renderer. The mirror and glass examples use a separate recursive ray tracer. During editing, the acceleration structure is rebuilt each frame. On the 60-frame hats smudge sequence, moving all 500k components, rebuilding, and rendering a $768\times512$ frame takes 10.0\,ms median (10.8\,ms maximum). These measurements show that carrying appearance with the components adds little cost to a full edit frame relative to a static render, without refitting the decoder.

\paragraph{Render times}
On one RTX 4090 with 500k lifted components, the glass-ball checkerboard scene (1024$\times$768, two bounces, point light) renders in a median of 51\,ms per frame with $2\times2$ supersampling, and in 26\,ms without the inserted ball. The Ambassadors mirror scene (1200$\times$900) takes 175\,ms with the mirror and 114\,ms without it. With one ray per pixel, the checkerboard scene takes 28\,ms with glass and 10\,ms without it. The Ambassadors scene takes 69\,ms with the mirror and 45\,ms without it. Times are CUDA-synchronized per frame and exclude image writing. These recursive ray-tracer timings, separate from 2D fitting rates, support interactive previews with moderate added cost from glass or the mirror.

\section{Additional applications}
We report super-resolution, stylization, geometric edits, and physical simulation with the fitted components.
\paragraph{Super-resolution}
For $8\times$ super-resolution, we train on a bicubic-downsampled low-resolution image and evaluate against its held-out high-resolution source. The 3000-iteration \method{} fits render at twice the low-resolution sampling rate, downsample with the known antialiased bicubic operator, and evaluate the loss at low resolution. Because the objective uses the known degradation operator, iterative back-projection~\cite{irani1991ibp} (IBP) is the relevant reference. Table~\ref{tab:sr} identifies which methods use this information. On the 20 images not used for recipe selection, \method{} retains a 0.32\,dB PSNR advantage over IBP. The 23-image mean in Table~\ref{tab:sr} includes the three recipe-selection images. In that table, \method{} leads in PSNR, SSIM, and LPIPS and beats IBP on every image. The MiraGe rerun matches its rasterizer support across output scales. Fitting the known degradation operator through the same local field lets its components be sampled on a finer grid without losing editability.

\begin{figure}[!t]
\centering
\includegraphics[width=\columnwidth,height=.62\textheight,keepaspectratio]{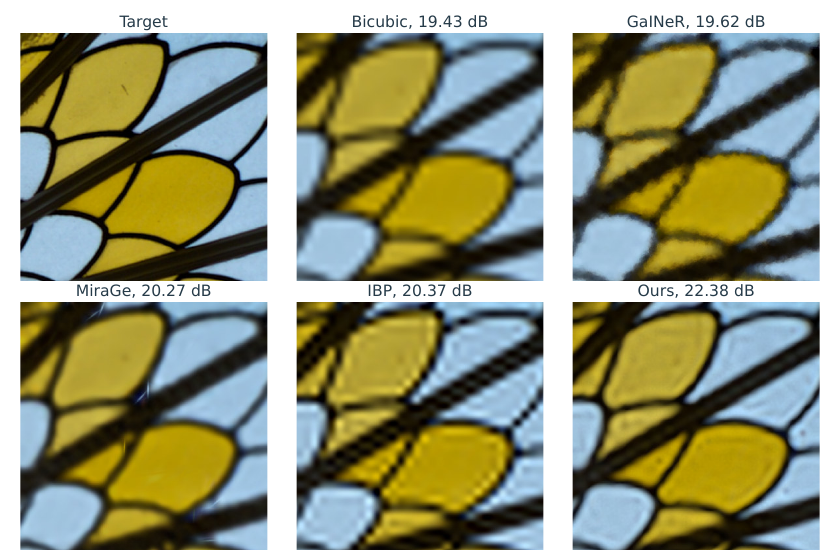}
\caption{$8\times$ super-resolution on DIV2K image 0821, the largest \method{} gain over IBP in our 23-image set. The ground-truth crop and five evaluated methods share the same output coordinates. Full-image PSNR appears above each reconstruction.}
\label{fig:supp_sr}
\end{figure}

\begin{table}[!htb]
\centering\small
\setlength{\tabcolsep}{3pt}
\caption{DIV2K $8\times$ super-resolution on 23 images. \emph{Operator} marks methods that use the bicubic degradation model. The groups have different information. Best reported values overall are bold.}
\label{tab:sr}
\begin{tabular}{lcccc}
\toprule
\textbf{Method} & \textbf{Operator} & \textbf{PSNR $\uparrow$} & \textbf{SSIM $\uparrow$} & \textbf{LPIPS $\downarrow$}\\
\midrule
Bicubic & no & 23.14 & .637 & .520\\
GaINeR & no & 23.16 & .636 & .520\\
MiraGe & no & 23.26 & .639 & .514\\
\midrule
IBP (20 steps) & yes & 23.55 & .648 & .501\\
\textbf{\method{} (ours)} & yes & \best{23.85} & \best{.658} & \best{.489}\\
\bottomrule
\end{tabular}
\end{table}

Fig.~\ref{fig:supp_sr} shows the largest gain over IBP, 2.01\,dB, among the evaluated images. All methods use the same crop, while labels report full-image PSNR. GaINeR and MiraGe do not use the bicubic degradation operator during optimization, so IBP is the more direct comparator for this experiment.

\begin{figure}[!tb]
\centering
\includegraphics[width=\columnwidth]{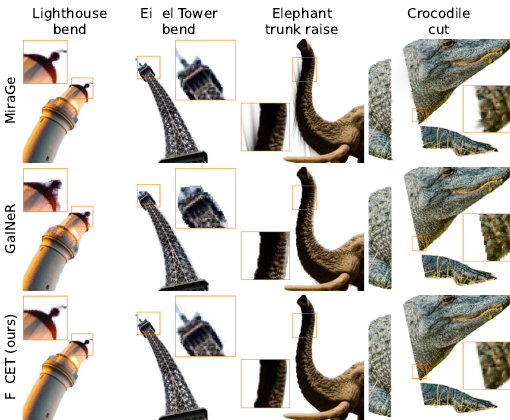}
\caption{Geometric edits with the same warp and object selection for all methods. Each column shows enlarged crops of the last frame, with MiraGe, GaINeR, and \method{} from top to bottom. $2$--$3\times$ insets enlarge the boxed edges, where GaINeR's binary mask leaves staircasing, light fringes, and a missing spire, while \method{} preserves smooth silhouettes. MiraGe leaves spikes along the raised trunk and ragged, blurred Gaussians at the crocodile cut.}
\label{fig:supp_edits}
\end{figure}

\paragraph{Geometric edits}
Fig.~\ref{fig:edits} shows two of the four geometric edits for MiraGe, GaINeR, and \method{} under the same warp and a shared object selection expanded by four pixels. Fig.~\ref{fig:supp_edits} shows all four edits.

MiraGe is fitted on the same RGBA inputs with its reconstruction-table setting (2D mirror camera, 500k initial Gaussians, 30k iterations) on a white background. A Gaussian moves when its projected center lies in the same 4-pixel-dilated object mask used for \method{}. Its center follows the warp, and its two in-plane axes are multiplied by the warp Jacobian at the center. The pipelines handle visibility differently: GaINeR derives a binary rendering mask from its points, while \method{} learns opacity from RGBA supervision. Consequently, an edge difference already present at the start of an edit cannot be assigned solely to ellipse transport. After removing two small detached alpha fragments from the lighthouse input, both methods were refitted with unchanged settings. The panels are arranged with one column per edit and one row per method.

\begin{figure}[!htb]
\centering
\includegraphics[width=\columnwidth,height=.62\textheight,keepaspectratio]{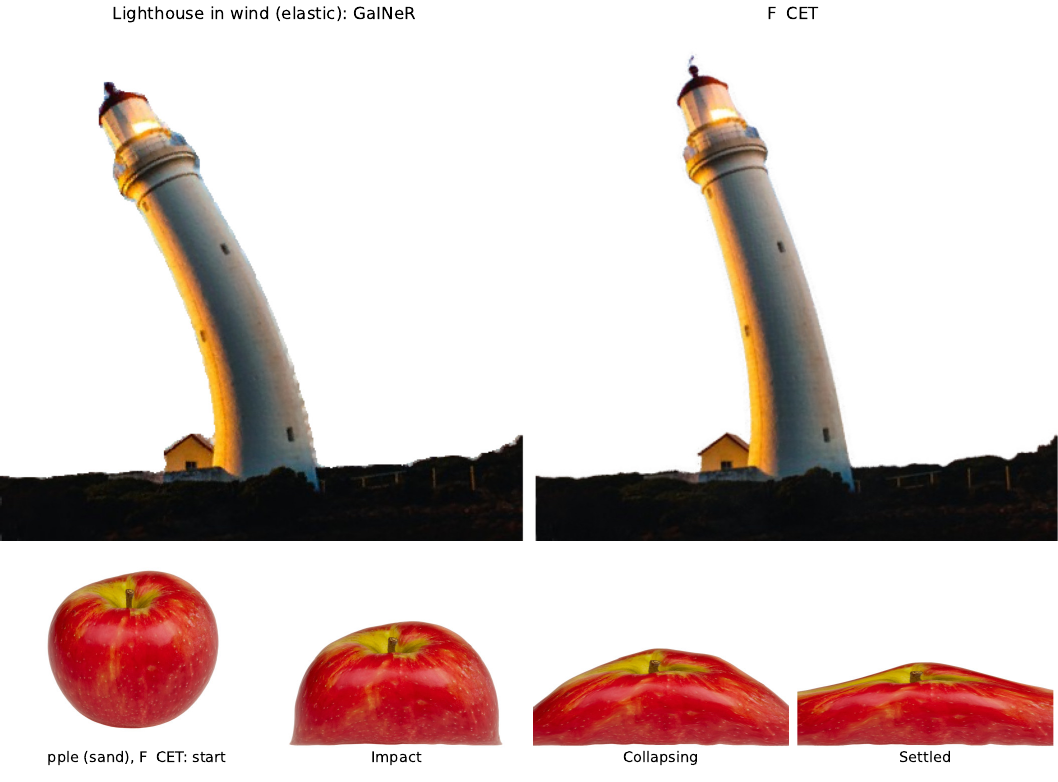}
\caption{MPM simulations. Top: lighthouse in wind (elastic) with matched solver settings, GaINeR (left) and \method{} (right). Bottom: an apple collapsing as sand with \method{} (start, impact, collapsing, settled).}
\label{fig:supp_physics}
\end{figure}

\paragraph{MPM simulations}
Fig.~\ref{fig:supp_physics} compares a lighthouse under matched MPM settings and shows an apple collapsing as sand. The latter breaks the source image's connectivity, illustrating that the component field can follow particle motion without a fixed image grid.

\begin{figure*}[!t]
\centering
\includegraphics[width=.94\textwidth,height=.59\textheight,keepaspectratio]{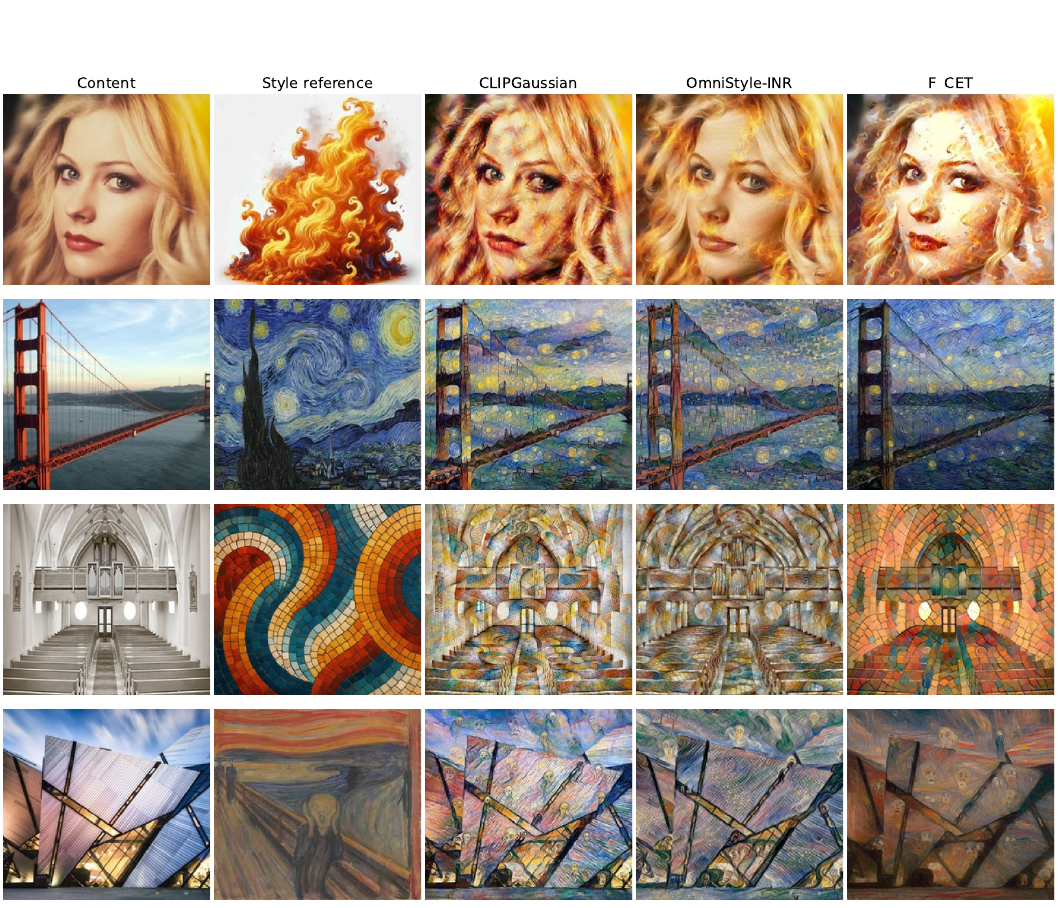}
\caption{Four examples from the OmniStyle-INR image-style set: content, reference style, CLIPGaussian~\cite{howil2025clipgaussian}, OmniStyle-INR~\cite{kajca2026omnistyle}, and our CLIP-and-Gram stylization result.}
\label{fig:supp_style}
\end{figure*}

\paragraph{Additional stylization measurements}
For stylization, we optimize appearance for a target style while holding geometry fixed. Four image-style and four text-style examples serve as demonstrations under a common CLIP-based evaluation, not a general ranking. The objective combines directional and patch CLIP~\cite{radford2021clip} style terms~\cite{kwon2022clipstyler} with a content term. Image-conditioned stylization additionally uses a Gram-matrix style term~\cite{gatys2016style}. On the OmniStyle-INR~\cite{kajca2026omnistyle} image-style set, \method{} has higher CLIP-based style similarity (71.7) than OmniStyle-INR (70.0) and CLIPGaussian~\cite{howil2025clipgaussian} (66.4) under a common evaluation. On the separate text-style set, it nearly matches CLIPGaussian in style similarity (26.2 versus 26.3) while retaining more content (SSIM 0.50 versus 0.45). These small-set results show that optimizing appearance alone can restyle a fit while leaving its geometry available for later deformation or lifting.

Fig.~\ref{fig:supp_style} shows all four image-style examples. Content, style references, and comparator renders come from the OmniStyle-INR assets~\cite{kajca2026omnistyle}. The \method{} column uses CLIP-and-Gram optimization. Across these examples, the fitted appearance changes while the same component geometry remains movable.

\begin{figure}[!htb]
\centering
\includegraphics[width=\columnwidth,height=.62\textheight,keepaspectratio]{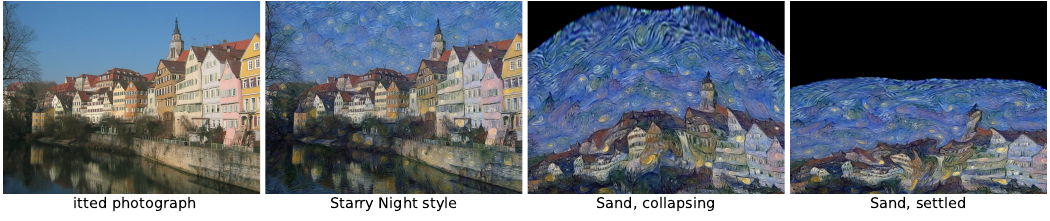}
\caption{Stylization and simulation combined: the fitted photograph, its Starry Night stylization, and the same stylized field simulated as sand (collapsing, settled).}
\label{fig:supp_style_physics}
\end{figure}

\paragraph{Stylization followed by simulation}
Fig.~\ref{fig:supp_style_physics} follows one fitted field from stylization into sand simulation. The style remains on the moving components because the simulation transports their local frames.

\begin{figure*}[!t]
\centering
\includegraphics[width=.9\textwidth,height=.62\textheight,keepaspectratio]{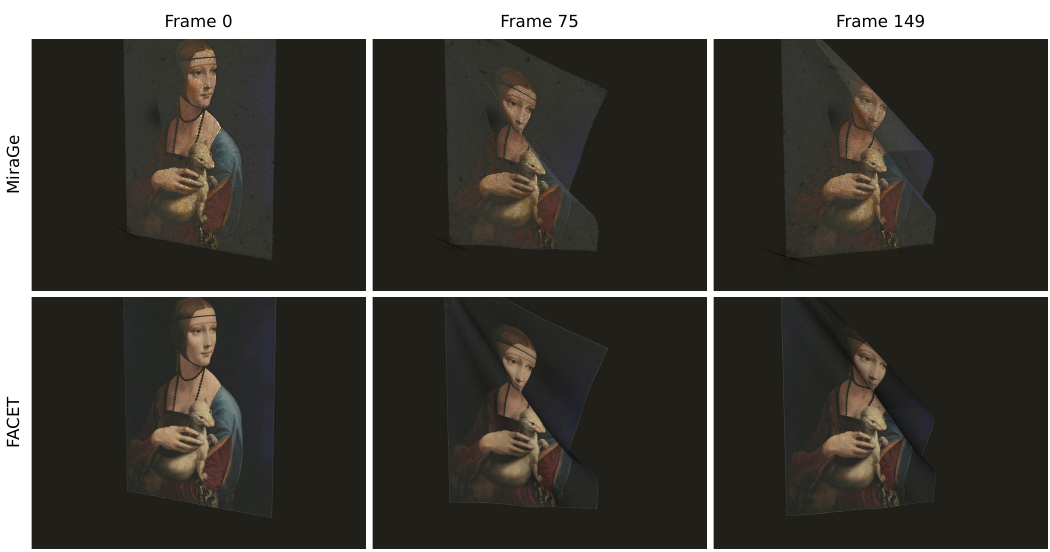}
\caption{Rendered \emph{Lady with an Ermine} on the same pinned-flag mesh and simulation. MiraGe in single-camera mode (top) has a 62.8\,dB frontal fit, but produces speckle, gray regions, and boundary artifacts in the shared oblique camera. \method{} (bottom) retains cleaner appearance in these views. Both sequences hold fitted appearance fixed and show the same three mesh states.}
\label{fig:supp_mirage_cloth}
\end{figure*}

\paragraph{Matched cloth comparison}
Fig.~\ref{fig:cloth} uses identical crops for the two methods at each displayed mesh state.
Both representations follow the same mesh and Warp cloth sequence, cameras, $2\times$ supersampling, and two-sided Lambert shading. MiraGe's cage deformation matches the \method{} cage to numerical precision. The sphere/background is taken from a render with the print transparent and is shared by both methods. MiraGe's full SH colors are evaluated along its training rays, corresponding to the fixed view-direction input used by \method{}. For each cloth scene, we select the MiraGe mode most favorable over the full sequence, with training-direction colors held fixed during the oblique-view renders.

For \emph{Great Wave}, MiraGe's mirror-camera mode gives its best frontal fit (62.3\,dB, versus 51.2 and 51.3\,dB for the one-camera and amorphous modes) and the fewest artifacts over the sequence, so the main figure uses it. Oblique views still reveal elongated boundary splats before motion and washed-out fold colors. The amorphous mode gives depth streaks, while the one-camera mode gives color speckle and boundary splats. Because mesh motion and rendering conditions are shared, these differences are consistent with how each fitted representation behaves away from the frontal view.

\paragraph{Lady with an Ermine}
A second image is fitted and attached to a flag pinned at its left edge (Fig.~\ref{fig:supp_mirage_cloth}). The figure compares registered undeformed and folded views under the same motion.

MiraGe's single-camera fit reaches 62.8\,dB frontal PSNR. The changed camera reveals speckle on the face, hands, and ermine, a gray background and back side, and boundary spikes. Its mirror-camera fit reaches 65.1\,dB frontally but shows stronger oblique-view speckle and smearing, with a mirrored back layer visible on the fold. We show the more favorable single-camera mode. Since these artifacts appear in the registered undeformed view, the difference cannot be assigned to cloth motion alone.

\begin{figure}[!htb]
\centering
\includegraphics[width=\columnwidth,height=.62\textheight,keepaspectratio]{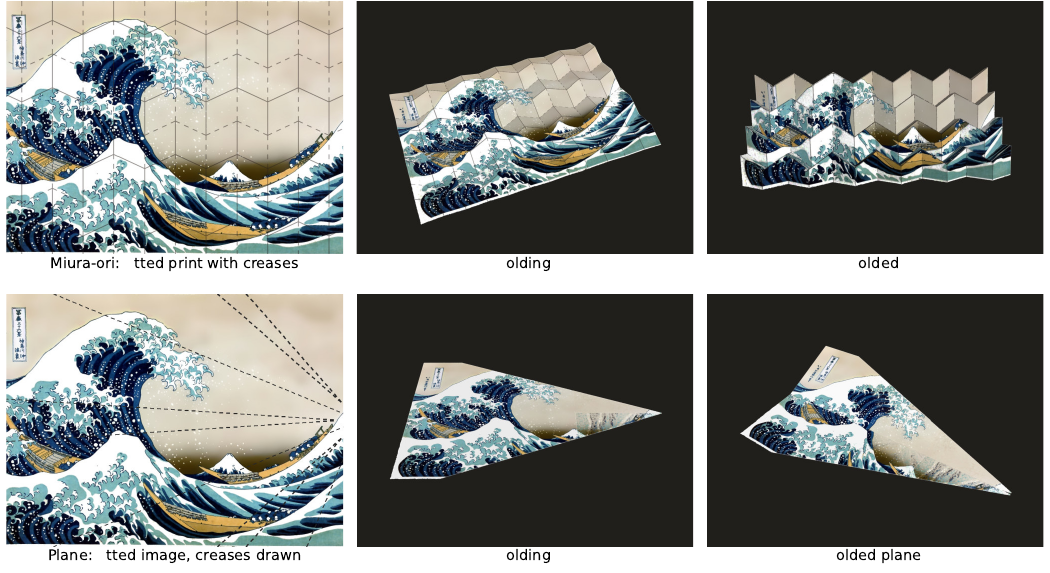}
\caption{Mesh-driven origami on fitted images. Top: Miura-ori, from the fitted print with its crease pattern to the folded sheet. Bottom: a paper plane, from the fitted image with the plane's creases drawn to the folded plane.}
\label{fig:supp_origami_gallery}
\end{figure}

\paragraph{Origami}
The fitted \emph{Great Wave} appearance can also follow a prescribed origami mesh fold (Fig.~\ref{fig:supp_origami_gallery}), without appearance refitting. This is a qualitative demonstration of the same cage binding used for cloth.

\twocolumn[{
\section{Gallery}
Fig.~\ref{fig:supp_teaser} collects further single-image scene assets, including a lifted room with an inserted glass sphere and Miura-ori folding. Panel (b) shows the original and styled T\"ubingen images. The \emph{Ambassadors} mirror scene appears in Fig.~\ref{fig:supp_effects}(e,f).

\begin{center}
\centering
\includegraphics[width=.95\textwidth,height=.62\textheight,keepaspectratio]{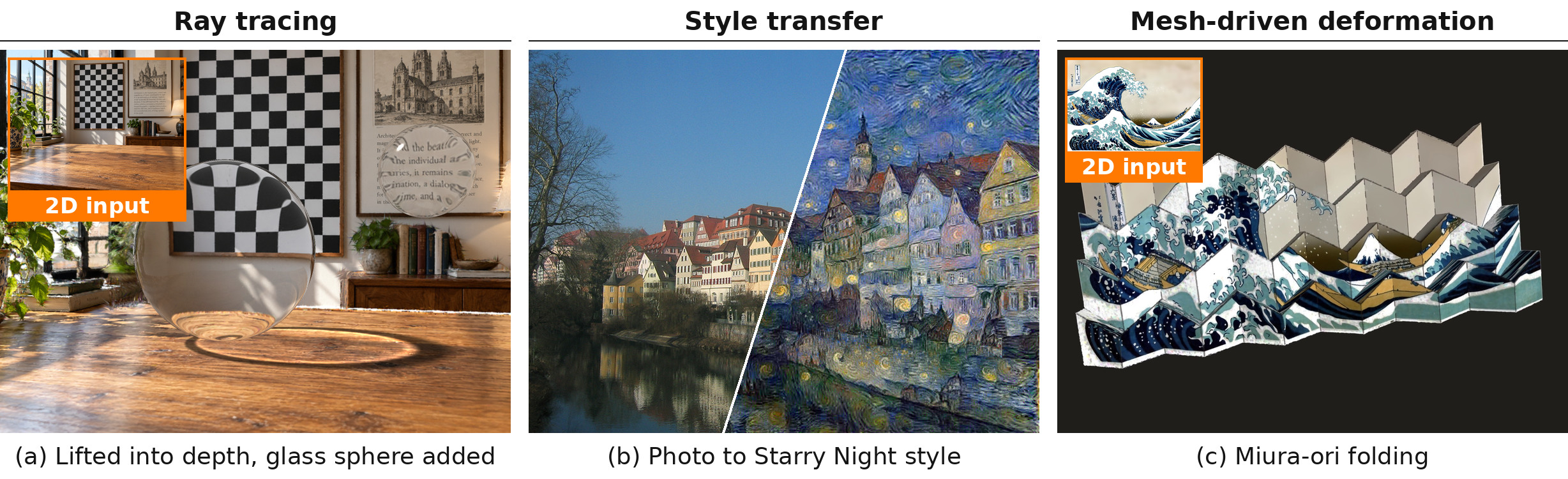}
\captionof{figure}{Each panel starts from a fit of one 2D image: (a) a checkerboard room lifted into depth with an inserted glass sphere, (b) a photograph restyled as \emph{Starry Night}, and (c) the fitted \emph{Great Wave} following a Miura-ori fold. Geometric operations retain fitted appearance. Stylization optimizes it.}
\label{fig:supp_teaser}
\end{center}

Fig.~\ref{fig:supp_effects} shows further optical effects on fitted and lifted images.

\begin{center}
\centering
\includegraphics[width=.95\textwidth,height=.62\textheight,keepaspectratio]{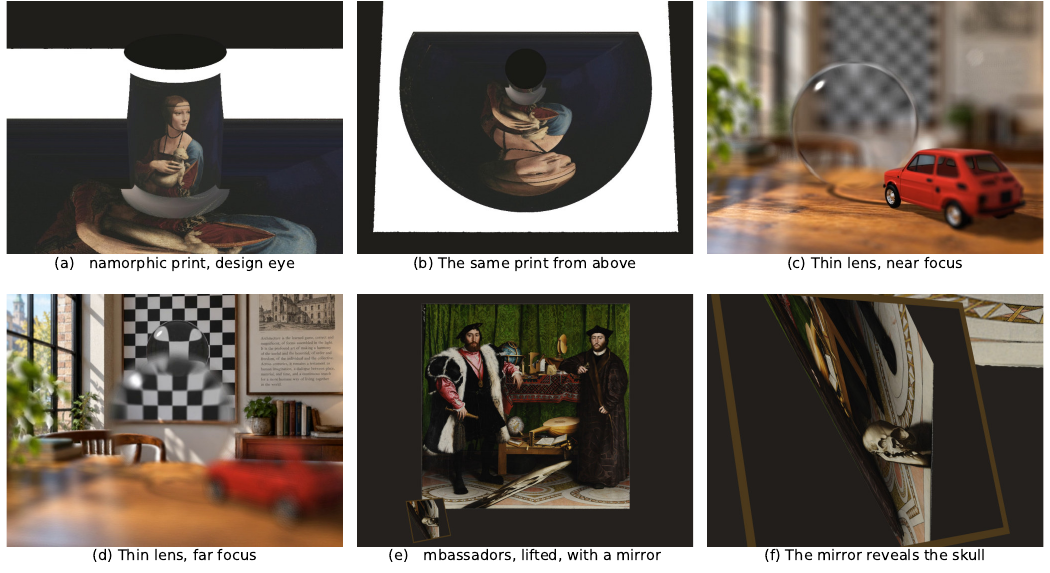}
\captionof{figure}{Additional ray-traced effects on fitted and lifted images (rendered scenes): (a,b) cylindrical anamorphosis seen from the design eye and from above, (c,d) a ray-traced thin lens focused near and far, (e,f) The Ambassadors lifted with a planar mirror revealing the skull.}
\label{fig:supp_effects}
\end{center}
\vspace{1ex}
}]

\FloatBarrier

\end{document}